\pdfoutput=1

\documentclass[11pt]{article}

\usepackage[]{acl}

\usepackage{times}
\usepackage{latexsym}
\usepackage{amsmath} 

\usepackage[T1]{fontenc}

\usepackage[utf8]{inputenc}

\usepackage{microtype}
\usepackage{algpseudocode}
\usepackage{multirow}
\usepackage{booktabs}
\usepackage{subfigure}
\usepackage{pifont}
\usepackage{CJK}

\usepackage{graphicx}
\usepackage{algorithm}
\usepackage{algorithmicx}
\usepackage{algpseudocode}

\renewcommand{\algorithmiccomment}[1]{\bgroup\hfill//~#1\egroup}

\title{Towards Generalized Open Information Extraction}

\author{
  Bowen Yu$^{1,2}$,
  Zhenyu Zhang$^{2,3}$,
  Jingyang Li$^{1}$,
  Haiyang Yu$^{1}$,\\ 
\bf
  Tingwen Liu$^{2,3}$\textsuperscript{\thanks{\quad Corresponding author.}},
  Jian Sun$^{1}$, 
  Yongbin Li$^{1}$\textsuperscript{\footnotemark[1]},
  Bin Wang$^{4}$\\
  $^1$ DAMO Academy, Alibaba Group \\
  $^2$Institute of Information Engineering, Chinese Academy of Sciences \\
 $^2$School of Cyber Security, University of Chinese Academy of Sciences \\
  $^3$Xiaomi AI Lab, Xiaomi Inc., Beijing, China\\
  {\tt \{yubowen.ybw,qiwei.ljy,yifei.yhy,shuide.lyb\}@alibaba-inc.com} \\
  \tt \{zhangzhenyu1996,liutingwen\}@iie.ac.cn wangbin11@xiaomi.com 
}

\begin{document}
\maketitle
\begin{abstract}

Open Information Extraction (OpenIE) facilitates the open-domain discovery of textual facts. 
However, the prevailing solutions evaluate OpenIE models on in-domain test sets aside from the training corpus, which certainly violates the initial task principle of domain-independence.
In this paper, we propose to advance OpenIE towards a more realistic scenario: generalizing over unseen target domains with different data distributions from the source training domains, termed Generalized OpenIE.
For this purpose, we first introduce GLOBE, a large-scale human-annotated multi-domain OpenIE benchmark, to examine the robustness of recent OpenIE models to domain shifts, and the relative performance degradation of up to 70\% implies the challenges of generalized OpenIE.
Then, we propose DragonIE, which explores a minimalist graph expression of textual fact: directed acyclic graph, to improve the OpenIE generalization.
Extensive experiments demonstrate that DragonIE beats the previous methods in both in-domain and out-of-domain settings by as much as 6.0\% in F1 score absolutely, but there is still ample room for improvement.

\end{abstract}


\section{Introduction}


Open Information Extraction (OpenIE) aims to mine open-domain facts indicating a semantic relation between a predicate phrase and its arguments from plain text~\cite{etzioni2008open}, without fixed relation vocabulary.
OpenIE developments have been demonstrated to benefit various domains and applications, such as knowledge base population~\cite{dong2014knowledge}, question answering~\cite{fader2014open}, and summarization~\cite{fan2019using}

Recently, OpenIE has seen remarkable advances.
Regarding different strategies for representing open fact, recent techniques with deep neural models can be subsumed under two categories, i.e., sequence-based and graph-based.
Sequence-based models predict the facts one by one in an auto-regressive fashion with iterative labeling or generation framework~\cite{cui2018neural,sun2018logician,kolluru-etal-2020-openie6,kolluru2020imojie}, which is the most classical solution in OpenIE.
Graph-based method formulates OpenIE as a maximal clique discovery problem based on the span-level text graph~\cite{yu2021maximal}, in which
the edge between two spans is defined as the combination of their roles in corresponding fact.
To the end, $O(m^2$) edges of $O(r^2$) types are constructed for a fact with $m$ spans of $r$ roles. 

Owning to the exquisite design, both sequence-based and graph-based models can identify complicated facts, thus constantly refreshing performance on benchmarks.
Nonetheless, it is still unexplored whether these models are sufficient for true open-domain extraction.
This doubt comes from that the training and test data in existing OpenIE benchmarks are generally independent and identically distributed, i.e., drawn from the same domain~\cite{stanovsky2018supervised,sun2018logician,gashteovski2019opiec}.
However, this assumption does not hold in practice.
Built on domain-independence~\cite{niklaus2018survey}, OpenIE models have to process diverse text, it is common to observe domain shifts among training and test data in applications.
Therefore, the performance on in-domain benchmarks may not exactly measure the generalization of out-of-domain extraction.




Starting from this concern, we carry out extensive experiments to investigate whether state-of-the-art OpenIE models preserve good performance on unseen target domains.
To provide a reliable benchmark, we publicize the first \emph{Generalized OpenIE} dataset containing 110,122 open facts annotated humanly on 20,899 sentences collected from 6 completely different domains.
We find out that, there are some noticeable semantic differences between open facts in different domains, posing challenges to the generalization of OpenIE models.
Because of domain shifts, in sequence-based models, the accuracy in each step prediction declines significantly, and the early errors are magnified later.
Similarly, in the graph-based model, the reduced edge prediction ability struggles to accurately connect $O(m^2)$ edges of $O(r^2)$ types especially when the span number $m$ and role number $r$ are both no small in complicated facts.
As a result, their F1 scores degrade as much as 70\% relatively (from 43\% to 13\%) when applied to unfamiliar domains, thus cannot work well in real-world extraction.

The above observations demonstrate full-fledged open-domain extraction still has a long way to go, and suggest a way for a more generalized OpenIE model: we should reduce the extraction complexity to lower the potential risk of prediction errors in domain shifts.
This is essentially the Occam’s Razor principle~\cite{rasmussen2000occam}: among all functions which fit the training data well, simpler functions are expected to generalize better.
 Therefore, we explore a minimalist expression of open fact: by sequentially connecting the boundary positions of all spans in the fact with their order in the text, each open fact can be simply modeled as a directed acyclic graph.
 Then OpenIE is equivalent to predicting the graph adjacency matrix and decoding facts from the directed graph.
 This idea leverages the sequential priors to reduce the complexity of function space (edge number and type) in the previous graph-based model from quadratic to linear, while avoiding auto-regressive extraction in   
  sequence-based models, thus improving generalization. 
 We implement it in DragonIE, a \textbf{D}i\textbf{r}ected \textbf{a}cyclic \textbf{g}raph based \textbf{o}pe\textbf{n} \textbf{I}nformation \textbf{E}xtractor.

 We perform extensive in-domain and out-of-domain experiments for OpenIE.
 On the previous commonly used in-domain evaluation, DragonIE outperforms the state-of-the-art method, with substantial gains of up to 3.6\% average F1 score, 3x speedup, and 5x convergence.
 Meantime, it reduces the number of edges by 66\% and the number of edge types by 88\% compared with the previous graph-based method.
 On our newly proposed out-of-domain benchmark, DragonIE further improves the performance gap to 6.0\%, and still exceeds the previous methods with only 10\% training data, showing better generalization.
 Detailed analysis shows that DragonIE can effectively represent overlapping, nested, discontinuous, and multiple facts despite its simplicity.
We also perform a qualitative analysis that summarizes typical extraction errors and outlines the future directions.

\section{Pilot Experiment}

To quantitatively evaluate the robustness of OpenIE model against domain shifts, we first propose a standard evaluation setup for generalized OpenIE.
Then, we conduct pilot experiments as well as empirical analyses in this section.

\begin{table}[t]\small
	\centering
	\resizebox{\linewidth}{!}{
	\begin{tabular}{lrrccc}
		\toprule
		  Datasets & \#Sents & \#Facts  & Human? & Shift? \\
		\midrule
		OIE2016~(\citeyear{stanovsky2016creating})  & 3,180 & 8,477  & \ding{55} & \ding{55} \\
		SAOKE~(\citeyear{sun2018logician}) & 46,930 & 166,370   & \ding{51} & \ding{55}\\
		CaRB~(\citeyear{bhardwaj-etal-2019-carb}) & 1,282 & 5,263  & \ding{51} &\ding{55}\\
		OpenIE4~(\citeyear{kolluru2020imojie}) & 92,774 & 190,661  & \ding{55} & \ding{55}\\
		LSOIE-wiki~(\citeyear{solawetz2021lsoie}) & 24,296 & 56,662  & \ding{55} & \ding{55}\\
		\midrule
		GLOBE (our) & 20,899 & 110,122 & \ding{51} & \ding{51}\\
		\bottomrule
	\end{tabular}}
		\caption{Comparison of representative OpenIE datasets. Human means the dataset is human-annotated rather than model-derived or converted from other corpus. Shift denotes the dataset supports the evaluation of OpenIE generalization performance with domain shift. }
	\label{tab.comp}
\end{table}

\subsection{Generalized OpenIE Evaluation Setup}
\label{sec:globe}
Given a sentence, OpenIE aims to output a set of facts in the form of (\emph{subject}, \emph{predicate}, \emph{object}$_1$, $\cdots$, \emph{object}$_n$), and all of them are stated explicitly in the text~\cite{yu2021maximal}.
As shown in Table~\ref{tab.comp}, Most existing OpenIE datasets assume that the training and test data are identically distributed without domain shift, which is certainly opposite to the task principle of domain independence.
To address this issue, we present GLOBE, a 
\textbf{G}enera\textbf{L}ized \textbf{O}penIE \textbf{BE}nchmark.
Firstly, sentences in GLOBE are collected from six distinct data sources, including insurance, education, finance, government, medicine, and news, which distinguishes GLOBE from existing datasets.
Then, GLOBE is annotated following the guidelines of SAOKE~\cite{sun2018logician}, the largest human-annotated OpenIE dataset collected from Baidu Encyclopedia. Thus they can combine to produce a complete training-test evaluation setup, comprehensively evaluating generalized OpenIE.
Specifically, the models are first trained on the SAOKE training set, and then the model with the best performance on the SAOKE dev set is selected to output results on GLOBE.
The annotation details and descriptive statistics of GLOBE are presented in Appendix~\ref{sec:apd_dataset}.

\begin{figure}
    \centering
    \includegraphics[width=\columnwidth]{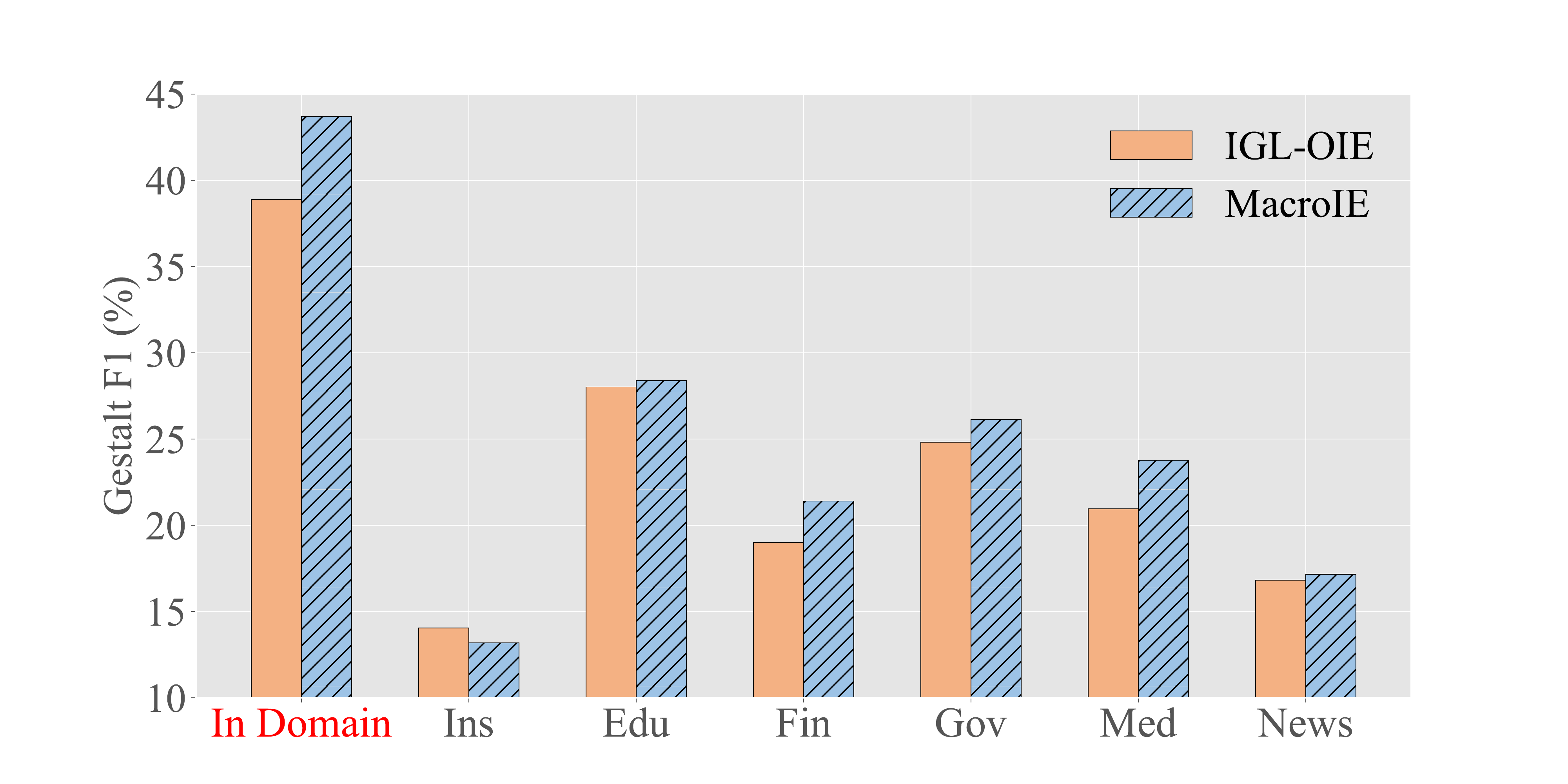}
    \caption{Gestalt F1 score comparasion on six out-of-domain test sets and the original in-domain test set.}
    \label{fig:empirical}
    \vspace{-0.1in}
\end{figure}

\subsection{Result Analysis}

We select the best-performing sequence model IGL-OIE~\cite{kolluru-etal-2020-openie6}, and graph model MacroIE~\cite{yu2021maximal}, for our pilot experiments. 
The evaluation metric is gestalt F1 score~\cite{yu2021maximal}.
Note that there are ore datasets and metrics in the main experiments (Section~\ref{sec:exp}).

Figure~\ref{fig:empirical} shows a detailed comparison across different domains and models on GLOBE. From the results we can see that: compared with the performances on SAOKE under in-domain setting, both the sequence-based and graph-based models encounter great performance drops on out-of-domain GLOBE, with a relative decline of 35\%-70\% in F1 score.
This indicates that the robustness of OpenIE model may be challenged in cross-domain generalization.
Intuitively, there are obvious differences in the topic and style of texts in different domains. 
For example, in the medical domain, subject and object are usually rare biological terminology, which is less covered in the limited general-domain training data.
Such a semantic shift degrades the prediction ability of the model fitted to the training set.
%
%
%
%


Exacerbating this issue further, modern OpenIE models often contain multiple prediction steps. 
Under domain shifts, every step is likely to go wrong, resulting in a collapse in the overall performance.
Specifically, sequence-based models predict facts auto-regressively, an mispredicted fact will directly affect the extraction of all the following facts.
The graph-based model requires $O(m^2)$ edges of $O(r^2)$ types for a fact with $m$ spans of $r$ roles. 
In GLOBE, the built graph contains an average of 28.5 edges with a total of 176 edge types for each open fact, and the wrong prediction of any edge may lead to the overall failure.
Thus, these methods are vulnerable to out-of-domain generalization.



\section{Methodology}
From the above observations, we know that recent OpenIE models are too complex to generalize. 
In this section, we propose a simplified expression of open fact: directed acyclic graph.
We start with the motivation of our new graph structure, then go through the implementation details.


\subsection{Motivation}

\begin{figure}
    \centering
    \includegraphics[width=\linewidth]{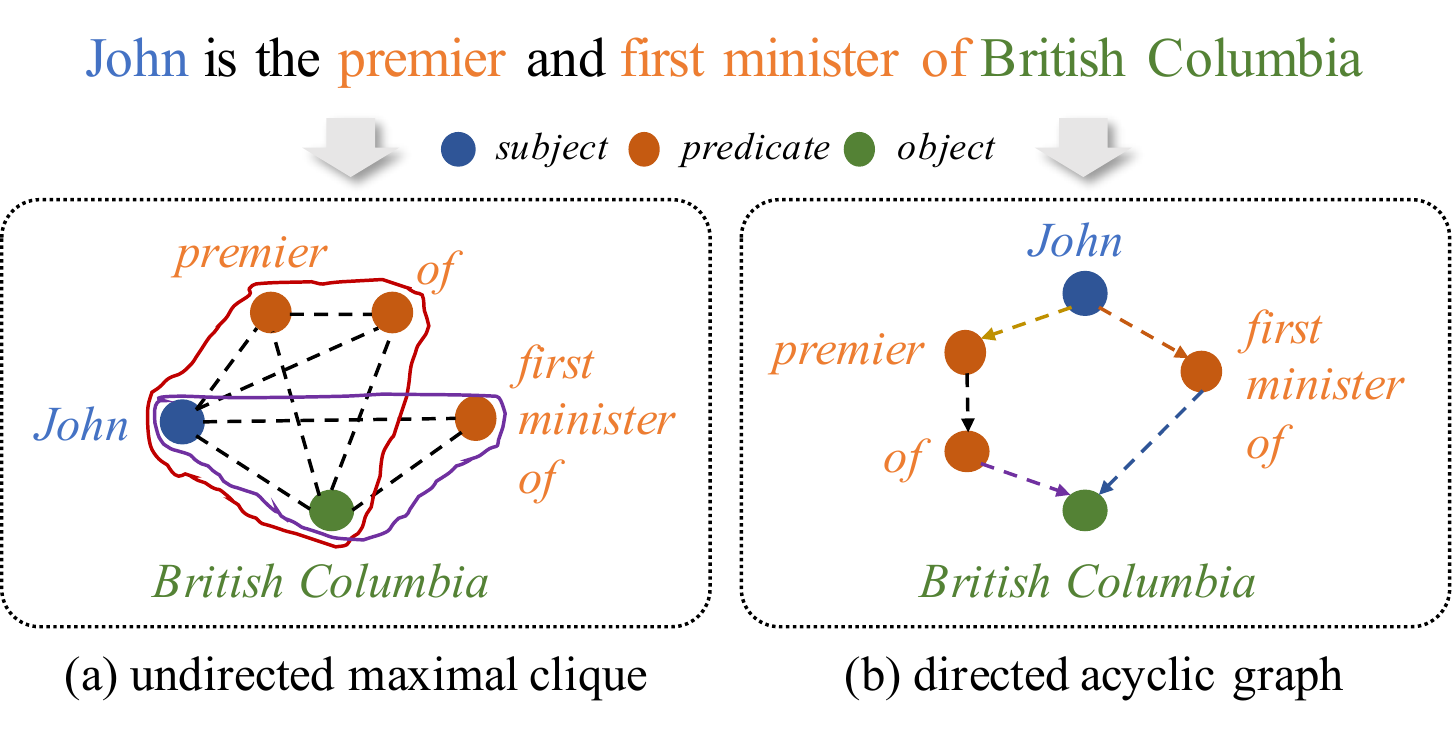}
    \caption{An example of representing open facts as an undirected maximal clique or a directed acyclic graph.}
    \label{fig:dag}
    \vspace{-0.1in}
\end{figure}

How to properly model open fact is the most important problem in OpenIE system design. 
The previous graph-based model treats spans belonging to one open fact as an undirected clique such that spans are pairwise connected with a combination of their roles as the edge type.
Whereas, as shown in Figure~\ref{fig:dag}, there is actually a natural reading order from left to right between spans in the text.
Such sequential prior means we can simply connect the edges between adjacent spans in the text to determine open facts.
In this way, the model no longer has to identify the pairwise relation between each span pair, which lessens the learning burden by reducing the edge numbers from $O(m^2)$ to $O(m)$.
Moreover, benefiting from the directed edge, we can assign the role of one connected span as the edge type, and recursively obtain the roles of all spans, thus greatly simplifying the edge type space from $O(r^2)$ to $O(r)$.
Meanwhile, the edges can be predicted in parallel, thus solving the cascade error in previous auto-regressive models.




\begin{figure*}
    \centering
    \includegraphics[width=0.988\linewidth]{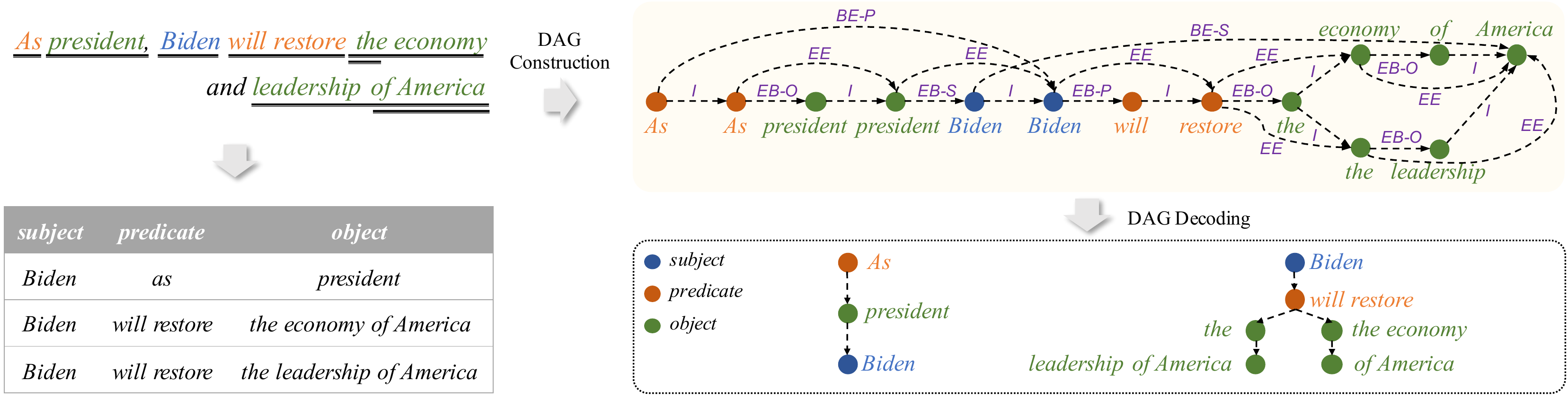}
    \caption{An overview of DargonIE. When building DAG, it enumerates each word pair and predict their edges. Thus, for spans with a single word, such as \emph{As}, there will be two vertexes refer to the beginning and ending words.}
    \label{fig:dag2}
    \vspace{-0.1in}
\end{figure*}




\subsection{Directed Acyclic Graph}
\label{sec:graph}
The above operation actually converts each input text to a directed acyclic graph (DAG).
In graph theory, a DAG consists of vertices and edges, with each edge directed from one vertex to another, such that following those directions will never form a closed loop. 
DAG can be topologically ordered, by arranging the vertices as a linear ordering with the edge directions.
This feature is consistent with what we want to combine span in the order that it appears in the text.
If we treat each continuous span involved in one fact asserted by the input text as a vertex in DAG, and connect oriented edges, from one vertex to another one that later appears in the text and belongs to the same fact.
Then in the simple case shown in Figure~\ref{fig:dag}, each directed path from root to leaf vertex represents an open fact.


Unfortunately, such an elegant paradigm is not suitable for all scenarios. 
When dealing with some complex cases like Figure~\ref{fig:dag2}, it encounters the following challenges:
(1) The granularity of text is word, while the granularity of open fact is span, so it is necessary to predict not only the relations between spans but also what is a span in the fact;
(2) Different spans may be overlapping and share some words, as the span
    \textit{of America} is enclosed in another span \textit{leadership of America} in the case of Figure~\ref{fig:dag2}.  
(3) Different facts may be overlapping and share some fact elements (either subject, predicate or object).
For example, \textit{Biden} acts as the subject in all the three facts and is not the root vertex.
Therefore, we cannot simply assume that each path in the DAG represents an open fact.

\textbf{DAG Construction.} 
These challenges prompt us to design the following three types of edges to avoid ambiguous extraction:
 (1) \emph{intra-span edge}: it connects the beginning and ending words of a span with a \verb|I| tag.
 (2) \emph{inter-span edge}: it connects the \textbf{E}nding word of a span and the \textbf{B}eginning/\textbf{E}nding word of the next span in the fact with a \verb|EB-X|/\verb|EE| tag, respectively, where 
 \verb|X| represents the role of the next span.
 Intuitively, each span can be uniquely identified by its two boundary words, and the double inter-span edge design helps distinguish overlapping spans.
 If we only connect the ending words of two spans, such as \textit{the} and \textit{America}, we cannot determine whether the subsequent span of \textit {the} is \textit{leadership of America} or \textit{of America}, because they have the same ending word, and it is the same with just using the \verb|EB-X| tag.
 (3) \emph{intra-fact edge}: it connects the \textbf{B}eginning word of the first span and the \textbf{E}nding word of the last span in a fact with a \verb|BE-X| tag to delimit the boundary of a fact. 
 In this way, even for overlapping facts, we can accurately judge the range of each fact within DAG.
 Because only the role of the subsequent span is indicated in the inter-span edge, the role of the first span in the fact is unknown, so we specify it in \verb|BE-X|.

 

 \textbf{DAG Decoding.} 
With the edge definition above, 
we first find all \verb|BE-X| edges to determine the beginning and ending words of target facts, and then traverse all paths between them, in which each path represents a fact.
During decoding each path, all the \verb|I| edges are utilized to determine the spans in the path, then we can judge the role of each span according to the \verb|EB-X| edge and distinguish overlapping spans with the \verb|EE| edge.
 Finally, spans in each path are combined according to their roles to output structured facts.
Besides, DAG can naturally identify discontinuous facts, where each element in open fact may contain multiple spans.
we can splice the spans of the same role in the order of the text to get the discontinuous element.
 In Section~\ref{sec:detailed_analysis}, we empirically conclude that our constructed DAG has been a minimalist expression of open fact: arbitrarily removing any edge will reduce the representation ability.
The Occam's Razor principle has stated that among all functions that have a good training set fit, the simplest one is likely to generalize better.
Thus DAG is expected to have great generalization in OpenIE.
  


\subsection{Architecture}
Therefore, OpenIE is transformed into how to build a desired DAG.
To this end, we propose DragonIE, a \textbf{D}i\textbf{r}ected \textbf{a}cyclic \textbf{g}raph based \textbf{o}pe\textbf{n} \textbf{I}nformation \textbf{E}xtractor.
Intuitively, the edges defined in DAG depict the relation between words in the text, so DragonIE enumerates all word pairs and makes parallel prediction\footnote{To clearly explore whether a more simplified graph structure can bring better generalization, we reuse the architecture of previous graph-based method~\cite{yu2021maximal} here.}:
\begin{align}
    \label{equ:encoder}
    \mathbf{h}_1,&\cdots,\mathbf{h}_n = {\rm Encoder}(w_1, \cdots, w_n), \\
    \label{equ:biaffine}    
    \mathbf{s}_{i,j} & = \mathbf{h}_i^\top \mathbf{U}\mathbf{h}_j + \mathbf{W}[\mathbf{h}_i;\mathbf{h}_j]+\mathbf{b},\\
    \mathbf{p}_{i,j}  &= {\rm Sigmoid}(\mathbf{s}_{i,j}).
\end{align}
It first maps each word $w_i$ into a $d$-dimensional contextual vector $\mathbf{h_i}\in \mathbf{R}^d$ with a basic encoder such as BERT~\cite{devlin2018bert}.
Then each ($\mathbf{h}_i$,$\mathbf{h}_j$) is fed to a pairwise score function, followed by a Sigmoid layer to yield the probability of each edge type $\mathbf{p}_{i,j}\in \mathbf{R}^c$~\cite{wang2020tplinker,wang2021discontinuous}.
During training, we optimize the parameters $\theta$ of DragonIE to minimize the cross-entropy loss:
\begin{align}
J(\theta)=&-\sum_{i=1}^n\sum_{j=i}^n\sum_{k=1}^c (\mathbf{y}_{i,j}[k]{\rm log}(\mathbf{p}_{i,j}[k]) \nonumber \\&+(1-\mathbf{y}_{i,j}[k]){\rm log}(1-\mathbf{p}_{i,j}[k])),
\end{align}
where $\mathbf{p}_{i,j}[k] \in [0,1]$ is the predicted probability of $(w_i,w_j)$ along the $k$-th edge type, and $\mathbf{y}_{i,j}[k]\in \{0,1\}$ is ground truth.
At inference, a threshold $\delta$ tuned on the dev set is applied to filter low confidence prediction and get the final edge labels.

\section{Experimental Setup}
\label{sec:exp}

\subsection{Datasets}
In our experiments, we evaluate the models on three datasets. 
(1) \textbf{SAOKE}~\cite{sun2018logician} is the largest human-annotated OpenIE dataset annotated from Baidu Encyclopedia. 
It contains 20k samples for training, 2k for validation, and 2k for testing.
Their division is independent and identically distributed so that SAOKE can be used as the standard dataset under the in-domain setting.
(2) \textbf{GLOBE} is the largest multi-domain OpenIE test set proposed in Section~\ref{sec:globe}. It follows the annotation scheme of SAOKE, but the domains are different, so it can effectively verify the performance of OpenIE models under the out-of-domain setting.
(3) \textbf{CarB}~\cite{bhardwaj-etal-2019-carb} is the first crowdsourced OpenIE dataset containing 1,282 sentences. 
Recently, it is widely used in testing models trained on OpenIE4~\cite{kolluru2020imojie}.
However, OpenIE4 is automatically-derived with great data noise, and the annotation scheme is inconsistent with CarB, so the results on CarB are relatively unreliable.

\subsection{Implementation Details}

We implement DragonIE by initializing the encoder parameters from BERT for English~\cite{devlin2018bert} and Chinese~\cite{cui-etal-2020-revisiting}.
DragonIE is optimized by BertAdam with a maximum sequence length of 200, an epoch number of 30, and a learning rate of 1e-5.
The threshold $\delta$ is selected from $[0.2, 0.4]$.
We select the model with best performance on validation set to output results on test set.
Hyper-parameters are selected based on the validation set, and all experiments are conducted on a single Tesla V100 GPU.

\subsection{Baselines and Evaluation metrics}

We employ recent neural models as strong baselines: sequential labeling (IGL-OIE~\cite{kolluru-etal-2020-openie6}), sequential generation (IMoJIE~\cite{kolluru2020imojie}), and graph-based (MacroIE~\cite{yu2021maximal}) models.
Following the convention~\cite{yu2021maximal},
we evaluate the performance with three most widely adopted metrics: CaRB-single~\cite{kolluru-etal-2020-openie6}, CaRB-multi~\cite{bhardwaj-etal-2019-carb} and Gestalt~\cite{sun2018logician}.
Each criterion produces three values: 
F1 score, the area under P-R curve (AUC), 
and the point in the P-R curve corresponding to the optimal F1 (Opt. F1).

\begin{table*}
\centering {\footnotesize
\begin{tabular}{lcccccccccc} 
\toprule       
Model $\downarrow$ - Metric $\rightarrow$ & \multicolumn{3}{c}{CaRB-single} & \multicolumn{3}{c}{CaRB-multi} & \multicolumn{3}{c}{Gestalt}
        \\ \cmidrule(r){2-4} \cmidrule(r){5-7} \cmidrule(r){8-10}
      & F1 & AUC & Opt.F1 & F1 & AUC & Opt.F1 & F1 & AUC & Opt.F1   \\
       
\midrule                         
IMoJIE~\cite{kolluru2020imojie}
& 36.6 & 22.6 & 37.0 & 38.7 & 25.4 & 39.5 & 36.4 & 22.5 & 37.3\\
IGL-OIE~\cite{kolluru-etal-2020-openie6}
& 37.6 & 22.8 & 38.4 & 39.3 & 25.5 & 40.6 & 37.1 & 23.6 & 38.4 \\
MacroIE~\cite{yu2021maximal}
& 41.2 & 24.5 & 41.5 & 42.7 & 27.8 & 43.7 & 42.8 & 27.2 & 43.7 \\
\midrule 
DragonIE (ours)
& \textbf{45.0} & \textbf{29.0} & \textbf{45.1} & \textbf{46.6} & \textbf{31.3} & \textbf{46.7} & \textbf{46.1} & \textbf{30.1} & \textbf{46.1} \\
\bottomrule
\end{tabular}
\caption{\textbf{In-domain Evaluation:} Main results on the in-domain benchmark SAOKE.}
\label{tab:saoke-table}
}
\end{table*}

\begin{table*}[!htb]
\centering {\footnotesize
\begin{tabular}{lcccccccccc} 
\toprule       
Model $\downarrow$ - Metric $\rightarrow$ & \multicolumn{3}{c}{CaRB-single} & \multicolumn{3}{c}{CaRB-multi} & \multicolumn{3}{c}{Gestalt}
        \\ \cmidrule(r){2-4} \cmidrule(r){5-7} \cmidrule(r){8-10}
      & F1 & AUC & Opt.F1 & F1 & AUC & Opt.F1 & F1 & AUC & Opt.F1   \\
       
\midrule                         
IGL-OIE~\cite{kolluru-etal-2020-openie6}
& 24.9 & 10.5 & 25.1 & 27.5 & 10.5 & 27.7 & 21.1 & 8.2 & 21.7 \\
MacroIE~\cite{yu2021maximal}
& 25.5 & 10.0 & 25.6 & 27.1 & 11.4 & 27.2 & 22.4 & 7.5 & 22.5 \\
\midrule 
DragonIE (ours)
& \textbf{30.9} & \textbf{15.1} & \textbf{31.0} & \textbf{33.3} & \textbf{17.5} & \textbf{33.5} & \textbf{28.6} & \textbf{13.1} & \textbf{28.7} \\
\bottomrule
\end{tabular}
\caption{\textbf{Out-of-domain Evaluation:} Main results on the out-of-domain benchmark GLOBE.
}
\label{tab:globe-table}
}
\end{table*}

\begin{table*}[!htb]
\centering {\footnotesize
\begin{tabular}{lcccccccccc} 
\toprule       
Model $\downarrow$ - Metric $\rightarrow$   & \multicolumn{3}{c}{CaRB-single} & \multicolumn{3}{c}{CaRB-multi} & \multicolumn{3}{c}{Gestalt}
        \\ \cmidrule(r){2-4} \cmidrule(r){5-7} \cmidrule(r){8-10}
      & F1 & AUC & Opt.F1 & F1 & AUC & Opt.F1 & F1 & AUC & Opt.F1   \\
       
\midrule                         
IGL-OIE~\cite{kolluru-etal-2020-openie6}
& 41.0 & 22.9 & 41.1 & 52.2 & 33.7 & 52.4 & 10.1 & 5.4 & 9.7 \\
MacroIE~\cite{yu2021maximal}
& 43.5 & 25.0 & 43.8 & 54.8 & 36.3 & 55.1 & 12.9 & 6.0 & 13.1 \\
\midrule 
DragonIE (ours)
& \textbf{43.9} & \textbf{25.3} & \textbf{44.1} & \textbf{55.1} & \textbf{36.4} & \textbf{55.1} & \textbf{13.6} & \textbf{6.3} & \textbf{13.7} \\
\bottomrule
\end{tabular}
\caption{\textbf{In-domain Evaluation:} Main results on  CaRB. The models are trained on the noisy OpenIE4 dataset. 
}
\label{tab:carb-table}
}
\end{table*}









\section{Experimental Results}

Our experiments aim to answer three questions:




\begin{description}
\setlength{\itemsep}{1pt}
\setlength{\parsep}{-1pt}
\setlength{\parskip}{-1pt}
\item[Q1] How does DragonIE compare to other methods in both in-domain and out-of-domain settings?
\item[Q2] Does DragonIE effectively handle complex extraction scenarios despite its simplicity?
\item[Q3] What causes the performance gap between out-of-domain  and in-domain OpenIE?
\vspace{-1.0em}
\end{description}

%
%



\begin{figure*}[t]
\centering
\subfigure[Complicated Extraction]{
\includegraphics[width=0.31\linewidth]{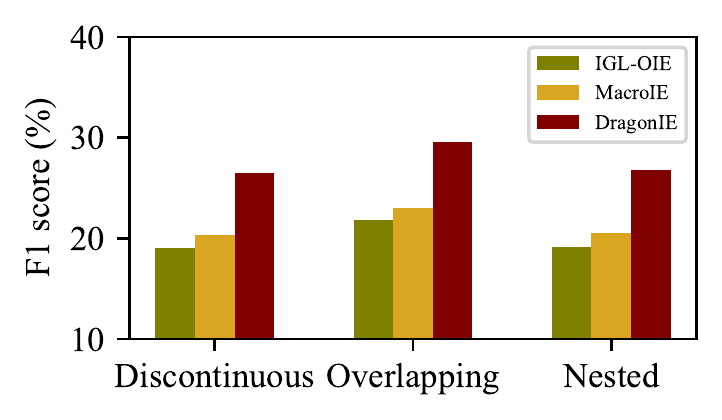}
\label{fig.stat.page}
}
\subfigure[Multiple Extraction]{
\includegraphics[width=0.31\linewidth]{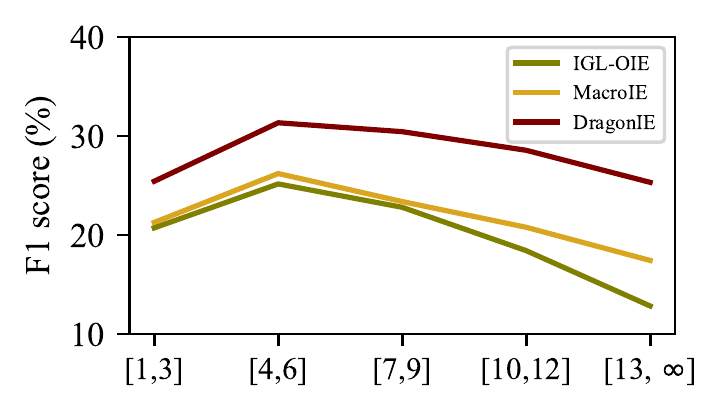}
\label{fig.stat.word}
}
\subfigure[Low-resource Extraction]{
\includegraphics[width=0.31\linewidth]{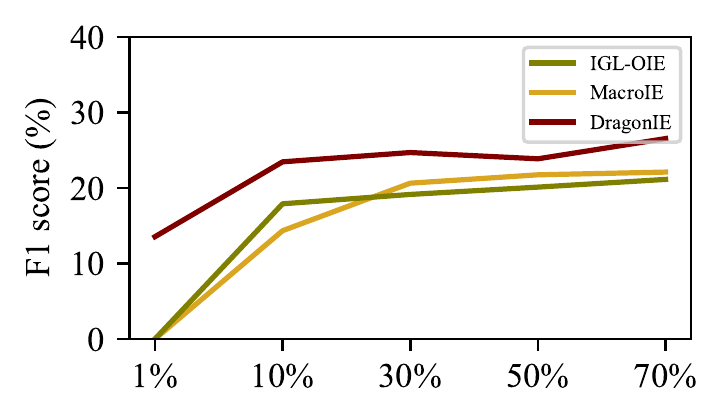}
\label{fig.stat.value}
}
\caption{Gestalt F1 scores on (a) complicated extraction, (b) multiple extraction, and (c) low-resource extraction. All the analyses are conducted on GLOBE. We also report the comparison results on SAOKE in Appendix B.3.}
\label{fig:complicated}
\vspace{-1.0em}
\end{figure*}

\subsection{Overall Performance (Q1)}
\label{sec:overall_performance}
Table~\ref{tab:saoke-table}-\ref{tab:carb-table} report the results of different models on all three datasets.
We can see that DragonIE establishes a new state-of-the-art for this task, and the improvement is statistically significant on the 5\% level for all datasets.
On the standard \textbf{in-domain} OpenIE benchmark SAOKE, DragonIE improves upon the previous best-performing model MacroIE in F1 score by absolute margins of 3.8, 3.9, and 3.3 points in CaRB-single, CaRB-multi, and Gestalt, respectively.
We use the models trained on SAOKE to get predictions on the \textbf{out-of-domain} benchmark GLOBE. 
Similar to in-domain experiments, 
DragonIE constantly achieves better results than existing methods, and the absolute gains are more impressive compared with the in-domain setting: from 3.6 to 6.0 F1 points on average, although there is still ample room for improvement (we will discuss it in Section~\ref{sec:error_analysis}).
The detailed comparison results in each domain of GLOBE are reported in Appendix~\ref{sec:detail_comp}.
Even for CaRB has much noise in the training data, our method still improves all evaluation metrics.
These observations verify that DragonIE has the flexibility to fact extraction, generalization to domain shift, and great robustness to data noise.
We believe this is because DragonIE explores a more concise and efficient OpenIE formulation, which avoids autoregressive prediction in previous sequence-based models, and simplifies the complexity of open fact in the graph-based model.
 In practice, to meet the complex extraction requirements, the maximal clique built by MacroIE for each open fact in SAOKE and GLOBE contains an average of 28.5 edges, with a total of 176 edge types, while DragonIE has an average of only 9.6 edges under 21 types.
 We provide a detailed edge space comparison in Appendix~\ref{sec:detail_edge}.
The simpler, the more essential, and the more effective.




Another advantage of simpler design is faster convergence and inference speed.
As shown in Table~\ref{tab:speed}, with the same hyper-parameters, DragonIE achieves the best results in 4 epochs, while MacroIE requires 20 epochs to reach the peak.
Moreover, DragonIE accelerates the testing time by 3 times.
While the decoding of MacroIE needs a time-consuming maximal clique discovery algorithm like Bron–Kerbosch~\cite{bron1973algorithm}, whose time complexity is O$(3^{n/3})$ for an $n$-vertex graph.
DragonIE avoids this issue, thus obtaining large speed improvement.

\begin{table}\small
\begin{center}
\begin{tabular}{lccc}
  \toprule
                    & MacroIE & DragonIE & Speedup \\
  \midrule
  Convergence (\emph{epoch})   & 4  & 20 & 5x  \\
  Testing (\emph{second})   & 136  & 409 & 3x \\
  \bottomrule
\end{tabular}
{\caption{Comparison in convergence and testing time on SAOKE, measured in epochs and seconds respectively.}
\label{tab:speed}}
\end{center}
\vspace{-2.0em}
\end{table}

\subsection{Detailed Analysis (Q2)}
\label{sec:detailed_analysis}

A potential concern is whether the better generalization of the simple DAG-based OpenIE formulation is at the expense of extracting complex facts, as simplicity usually leads to a reduction in representation capability.
To answer this question, we perform a fine-grained evaluation on GLOBE.
(1) We select the sentences containing discontinuous or overlapping or nested facts from GLOBE to form three complex test sets.
Here discontinuous means that at least one fact element in the sentence is not a continuous span, overlapping means that multiple facts in the sentence share at least one element, while nested means that different elements share some common spans.
These three patterns are the most common complex facts in OpenIE, and their distribution is detailed in Appendix~\ref{sec:dataset}.
(2) We validate DragonIE’s capability in extracting different numbers of open facts by splitting the sentences into five classes according to the fact count.
(3) We conduct low-resource experiments on five different partitions of the original SAOKE training sets (1/10/30/50/70\%).
As presented in Figure~\ref{fig:complicated}, DragonIE attains consistent gains in all classes across three settings, indicating that our model is more suitable for complicated scenarios than the baselines.
It is worth noting that when using 1\% of the training data, only DragonIE achieves a non-zero F1 score, and using 10\% of the training data can surpass the performance of baselines under the full data, indicating better generalization.

%
\begin{table}\small
\begin{center}
\begin{tabular}{lcc}
  \toprule
      Model $\downarrow$ - Benchmark $\rightarrow$                & GLOBE & SAOKE  \\
  \midrule
  DragonIE                & 28.6 & 45.8     \\
  \quad-- inner-span edge EE  & 26.1 & 45.1   \\
  \quad-- inter-fact edge BE-X  & 25.4 & 42.0     \\
  \quad-- Next span role labeling & 25.8 & 44.6     \\
  \bottomrule
\end{tabular}
{\caption{Ablation study of DragonIE. Numbers denote the corresponding Gestalt F1 scores.}
\label{tab:ablation}}
\end{center}
\vspace{-2.0em}
\end{table}

In addition, we conduct a set of ablation tests on the graph to verify that our DAG is already a minimalist expression of open fact.
Table~\ref{tab:ablation} shows that: 
(i) when only connecting the ending word of one span and the beginning word of the next span (EB-X) and removing the edge connected with the ending word of the next span (EE),  the F1 score drops by 1.6\% in average since it cannot accurately represent nested facts, as demonstrated in Section~\ref{sec:graph}; 
(II) Removing the intra-fact edge and treating each path from the root vertex to the leaf vertex on the DAG as a fact hurts the results by 3.5 F1 pts in average, which is difficult to extract overlapping facts; 
(III) Marking the role of the next span on edge instead of the combination of two-span roles brings a remarkable improvement (2.0\% averagely), since it effectively compresses the edge type space from $O(r^2)$ to $O(r)$.
Note that the intra-span edges cannot be ablated because they recognize spans.
On the whole, each edge in our built DAG is indispensable. 




\subsection{Qualitative Evaluation (Q3)}
\label{sec:error_analysis}

\begin{table}\small
\begin{center}
\begin{tabular}{lcc}
  \toprule
      Error $\downarrow$ - Benchmark $\rightarrow$               & GLOBE & SAOKE  \\
  \midrule
  Wrong Boundary               & 5  & 4     \\
  Wrong Extraction & 5 & 7 \\
  Uninformative Extraction & 13 & 10    \\
  Incomplete Extraction  & 12  & 2   \\
  Missing Extraction  & 26 & 17     \\
  \bottomrule
\end{tabular}
{\caption{Error analysis of DragonIE. We report the number of  false facts belonging to five major error classes on the analysis set (containing 100 gold facts) of in-domain and out-of-domain benchmarks.}
\label{tab:error}}
\end{center}
\vspace{-2.0em}
\end{table}


Although DragonIE achieves state-of-the-art results in all the benchmarks, there are still substantial differences between the out-of-domain and in-domain performance.
We compare the mistakes made by DragonIE with two analysis sets that sample from the test set of GLOBE and SAOKE, respectively, and summarize the error types.
The sampling strategy requires that the sentences in the analysis set contain 100 gold open facts.
Table~\ref{tab:error} reports five major error classes and the number of corresponding false facts on the two benchmarks.




\textbf{Wrong Boundary} is a too large or too small boundary for an element in an open fact.
\textbf{Wrong Extraction} describes an open fact that does not hold in the original sentence.
They are the least common error types in both settings, showing that our model can identify the correct span and fact across domains.
It would be interesting to see if introducing causal inference~\cite{nan2021uncovering}, or mutual information maximization~\cite{zhang2020unsupervised} to strengthen the correlation between facts and sentences, can improve the performance. 
\textbf{Uninformative Extraction} is widely present in the output of various domains, it usually does not provide information gain.
We think a promising improvement direction is applying an additional post-processing model to judge the informativeness of each open fact.
\textbf{Incomplete Extraction} omits critical information resulting in unclear fact semantics.
\textbf{Missing Extraction} is an outcome where the model fails to predict the open fact.
According to statistics from Table~\ref{tab:error}, these two types of errors are the root cause of the performance gap between in-domain and out-of-domain settings.
We believe the following research directions are worth following for them: 
(1) Pre-training models on a massive corpus with OpenIE-oriented self-supervised tasks to sufficiently capture domain-robust OpenIE exclusive features~\cite{lu2022unified};
(2) Leveraging the domain generalization techniques to learn the invariances across domains, i.g., meta learning~\cite{li2018learning,geng2019induction,zhao2022improving}, adversarial learning~\cite{li2018deep}, and contrastive learning~\cite{kim2021selfreg}.

\section{Related Work}

\noindent \textbf{OpenIE.} 
From rule-based systems and statistical methods~\cite{fader2011identifying, del2013clausie, gashteovski2017minie}, to neural models~\cite{cui2018neural,stanovsky2018supervised,roy2019supervising}, OpenIE research has experienced three technological evolutions in the past decade.
Each evolution brings a more expressive architecture, and meantime requiring much more training data.
To this day, the best-performing OpenIE model either predicts open facts in the sentence auto-regressively~\cite{kolluru-etal-2020-openie6,kolluru2020imojie}, or represents each open fact as a maximal clique on the graph with quadratic edge numbers and types~\cite{yu2021maximal}.
Such trends pose two potentially challenges:
(1) The popular evaluation protocol mainly operates with the i.i.d. assumption, i.e., the training domain is the same as the test domain~\cite{stanovsky2018supervised,sun2018logician,gashteovski2019opiec,yu2020joint,zhang2022layout}, which is contrary to the domain-independent discovery objective of OpenIE~\cite{niklaus2018survey}.
Although the existing studies have achieved surprising performance under i.i.d. evaluation, their generalization for true open extraction has not been evaluated.
Some works try to use OpenIE4~\cite{kolluru2020imojie} to train the model and verify it on CarB~\cite{bhardwaj-etal-2019-carb}, but the noise annotation of OpenIE4 and the different annotation standards of the two datasets make the evaluation results unreliable.
(2) As revealed by our preliminary experiments, recent OpenIE models always encounter great performance drops in the out-of-domain setting.
Their complex auto-regressive prediction process and graph structure may overfit the training data specifics, resulting in unsatisfactory cross-domain generalization.
In this paper, we present the first systematic study to examine how robust OpenIE methods are when trained and tested on different datasets (domains), and further propose a minimalist expression of open fact to implicitly improve the generalization behavior.

\noindent \textbf{Domain Generalization.} The main goal of domain generalization is to learn a domain-invariant representation from multiple source domains so that a model can generalize well across unseen target domains~\cite{kim2021selfreg,mi2021towards}.
Recent advances mainly focus on three aspects: data augmentation, model design, and robust training.
Augmenting the dataset with transformations such as mix-up~\cite{zhang2020does} improves generalization~\cite{pandey2021generalization}.
A simplified model design mines the task essence to resist domain shifts~\cite{ghosh2021network}.
Robust training methods hope to optimize a shared feature space,i.e, by minimizing maximum mean discrepancy~\cite{tzeng2014deep}, transformed feature distribution distance~\cite{muandet2013domain}, or covariances~\cite{sun2016deep}.
This paper primarily explores generalized OpenIE from the perspective of model design. 
How to combine data augmentation and robust training to further improve the generalization will be our future work.

\section{Conclusion}
In this paper, we lay out and study generalized OpenIE for the first time. 
We release GLOBE, a large-scale, high-quality, multi-domain benchmark with 110,122 open facts, to evaluate the generalization of OpenIE models. 
Furthermore, we explore the minimalist graph expression of open fact:  directed acyclic graph, to reduce the extraction complexity and improve the generalization behavior.
Experimental results show that our proposed method outperforms state-of-the-art baselines in both in-domain and out-of-domain settings.
This work is a starting point towards building more practical OpenIE models with stronger generalization, and we also present fine-grained analyses which point out promising avenues for further improvement.

\section{Limitations}
While this work has made some progress towards generalized OpenIE, it still has some limitations.
First, to produce a complete training-test evaluation setup with the largest human-annotated OpenIE dataset SAOKE, our annotated GLOBE benchmark is in Chinese.
We speculate that the same conclusions can be observed in other languages, and leave this for future work.
Second, although the proposed DragonIE method greatly exceeds the baselines, there is still a significant performance degradation under the out-of-domain setting compared with the in-domain setting. 
We will continue to work to narrow the performance gap.
\bibliography{new_anthology}

\begin{thebibliography}{42}
\expandafter\ifx\csname natexlab\endcsname\relax\def\natexlab#1{#1}\fi

\bibitem[{Bhardwaj et~al.(2019)Bhardwaj, Aggarwal, and
  Mausam}]{bhardwaj-etal-2019-carb}
Sangnie Bhardwaj, Samarth Aggarwal, and Mausam Mausam. 2019.
\newblock \href {https://doi.org/10.18653/v1/D19-1651} {{C}a{RB}: A
  crowdsourced benchmark for open {IE}}.
\newblock In \emph{Proceedings of the 2019 Conference on Empirical Methods in
  Natural Language Processing and the 9th International Joint Conference on
  Natural Language Processing (EMNLP-IJCNLP)}, pages 6262--6267, Hong Kong,
  China. Association for Computational Linguistics.

\bibitem[{Bron and Kerbosch(1973)}]{bron1973algorithm}
Coen Bron and Joep Kerbosch. 1973.
\newblock Algorithm 457: finding all cliques of an undirected graph.
\newblock \emph{Communications of the ACM}.

\bibitem[{Corro and Gemulla(2013)}]{del2013clausie}
Luciano~Del Corro and Rainer Gemulla. 2013.
\newblock \href {https://doi.org/10.1145/2488388.2488420} {Clausie:
  clause-based open information extraction}.
\newblock In \emph{22nd International World Wide Web Conference, {WWW} '13, Rio
  de Janeiro, Brazil, May 13-17, 2013}, pages 355--366. International World
  Wide Web Conferences Steering Committee / {ACM}.

\bibitem[{Cui et~al.(2018)Cui, Wei, and Zhou}]{cui2018neural}
Lei Cui, Furu Wei, and Ming Zhou. 2018.
\newblock \href {https://doi.org/10.18653/v1/P18-2065} {Neural open information
  extraction}.
\newblock In \emph{Proceedings of the 56th Annual Meeting of the Association
  for Computational Linguistics}, pages 407--413, Melbourne, Australia.
  Association for Computational Linguistics.

\bibitem[{Cui et~al.(2020)Cui, Che, Liu, Qin, Wang, and
  Hu}]{cui-etal-2020-revisiting}
Yiming Cui, Wanxiang Che, Ting Liu, Bing Qin, Shijin Wang, and Guoping Hu.
  2020.
\newblock \href {https://www.aclweb.org/anthology/2020.findings-emnlp.58}
  {Revisiting pre-trained models for {C}hinese natural language processing}.
\newblock In \emph{Proceedings of the 2020 Conference on Empirical Methods in
  Natural Language Processing: Findings}, pages 657--668, Online. Association
  for Computational Linguistics.

\bibitem[{Devlin et~al.(2019)Devlin, Chang, Lee, and
  Toutanova}]{devlin2018bert}
Jacob Devlin, Ming-Wei Chang, Kenton Lee, and Kristina Toutanova. 2019.
\newblock \href {https://doi.org/10.18653/v1/N19-1423} {{BERT}: Pre-training of
  deep bidirectional transformers for language understanding}.
\newblock In \emph{Proceedings of the 2019 Conference of the North {A}merican
  Chapter of the Association for Computational Linguistics: Human Language
  Technologies}, pages 4171--4186, Minneapolis, Minnesota. Association for
  Computational Linguistics.

\bibitem[{Dong et~al.(2014)Dong, Gabrilovich, Heitz, Horn, Lao, Murphy,
  Strohmann, Sun, and Zhang}]{dong2014knowledge}
Xin Dong, Evgeniy Gabrilovich, Geremy Heitz, Wilko Horn, Ni~Lao, Kevin Murphy,
  Thomas Strohmann, Shaohua Sun, and Wei Zhang. 2014.
\newblock \href {https://doi.org/10.1145/2623330.2623623} {Knowledge vault: a
  web-scale approach to probabilistic knowledge fusion}.
\newblock In \emph{The 20th {ACM} {SIGKDD} International Conference on
  Knowledge Discovery and Data Mining, {KDD} '14, New York, NY, {USA} - August
  24 - 27, 2014}, pages 601--610. {ACM}.

\bibitem[{Etzioni et~al.(2008)Etzioni, Banko, Soderland, and
  Weld}]{etzioni2008open}
Oren Etzioni, Michele Banko, Stephen Soderland, and Daniel~S Weld. 2008.
\newblock Open information extraction from the web.
\newblock \emph{Communications of the ACM}, 51(12):68--74.

\bibitem[{Fader et~al.(2011)Fader, Soderland, and
  Etzioni}]{fader2011identifying}
Anthony Fader, Stephen Soderland, and Oren Etzioni. 2011.
\newblock \href {https://aclanthology.org/D11-1142} {Identifying relations for
  open information extraction}.
\newblock In \emph{Proceedings of the 2011 Conference on Empirical Methods in
  Natural Language Processing}, pages 1535--1545, Edinburgh, Scotland, UK.
  Association for Computational Linguistics.

\bibitem[{Fader et~al.(2014)Fader, Zettlemoyer, and Etzioni}]{fader2014open}
Anthony Fader, Luke Zettlemoyer, and Oren Etzioni. 2014.
\newblock \href {https://doi.org/10.1145/2623330.2623677} {Open question
  answering over curated and extracted knowledge bases}.
\newblock In \emph{The 20th {ACM} {SIGKDD} International Conference on
  Knowledge Discovery and Data Mining, {KDD} '14, New York, NY, {USA} - August
  24 - 27, 2014}, pages 1156--1165. {ACM}.

\bibitem[{Fan et~al.(2019)Fan, Gardent, Braud, and Bordes}]{fan2019using}
Angela Fan, Claire Gardent, Chlo{\'e} Braud, and Antoine Bordes. 2019.
\newblock \href {https://doi.org/10.18653/v1/D19-1428} {Using local knowledge
  graph construction to scale {S}eq2{S}eq models to multi-document inputs}.
\newblock In \emph{Proceedings of the 2019 Conference on Empirical Methods in
  Natural Language Processing and the 9th International Joint Conference on
  Natural Language Processing (EMNLP-IJCNLP)}, pages 4186--4196, Hong Kong,
  China. Association for Computational Linguistics.

\bibitem[{Gashteovski et~al.(2017)Gashteovski, Gemulla, and del
  Corro}]{gashteovski2017minie}
Kiril Gashteovski, Rainer Gemulla, and Luciano del Corro. 2017.
\newblock \href {https://doi.org/10.18653/v1/D17-1278} {{M}in{IE}: Minimizing
  facts in open information extraction}.
\newblock In \emph{Proceedings of the 2017 Conference on Empirical Methods in
  Natural Language Processing}, pages 2630--2640, Denmark. Association for
  Computational Linguistics.

\bibitem[{Gashteovski et~al.(2019)Gashteovski, Wanner, Hertling, Broscheit, and
  Gemulla}]{gashteovski2019opiec}
Kiril Gashteovski, Sebastian Wanner, Sven Hertling, Samuel Broscheit, and
  Rainer Gemulla. 2019.
\newblock Opiec: An open information extraction corpus.
\newblock In \emph{AKBC}.

\bibitem[{Geng et~al.(2019)Geng, Li, Li, Zhu, Jian, and
  Sun}]{geng2019induction}
Ruiying Geng, Binhua Li, Yongbin Li, Xiaodan Zhu, Ping Jian, and Jian Sun.
  2019.
\newblock Induction networks for few-shot text classification.
\newblock In \emph{Proceedings of the 2019 Conference on Empirical Methods in
  Natural Language Processing and the 9th International Joint Conference on
  Natural Language Processing (EMNLP-IJCNLP)}, pages 3904--3913.

\bibitem[{Ghosh and Motani(2021)}]{ghosh2021network}
Rohan Ghosh and Mehul Motani. 2021.
\newblock Network-to-network regularization: Enforcing occam's razor to improve
  generalization.
\newblock \emph{Advances in Neural Information Processing Systems}, 34.

\bibitem[{Kim et~al.(2021)Kim, Yoo, Park, Kim, and Lee}]{kim2021selfreg}
Daehee Kim, Youngjun Yoo, Seunghyun Park, Jinkyu Kim, and Jaekoo Lee. 2021.
\newblock Selfreg: Self-supervised contrastive regularization for domain
  generalization.
\newblock In \emph{Proceedings of the IEEE/CVF International Conference on
  Computer Vision}, pages 9619--9628.

\bibitem[{Kolluru et~al.(2020{\natexlab{a}})Kolluru, Adlakha, Aggarwal,
  {Mausam}, and Chakrabarti}]{kolluru-etal-2020-openie6}
Keshav Kolluru, Vaibhav Adlakha, Samarth Aggarwal, {Mausam}, and Soumen
  Chakrabarti. 2020{\natexlab{a}}.
\newblock \href {https://doi.org/10.18653/v1/2020.emnlp-main.306} {{O}pen{IE}6:
  {I}terative {G}rid {L}abeling and {C}oordination {A}nalysis for {O}pen
  {I}nformation {E}xtraction}.
\newblock In \emph{Proceedings of the 2020 Conference on Empirical Methods in
  Natural Language Processing (EMNLP)}, pages 3748--3761, Online. Association
  for Computational Linguistics.

\bibitem[{Kolluru et~al.(2020{\natexlab{b}})Kolluru, Aggarwal, Rathore,
  {Mausam}, and Chakrabarti}]{kolluru2020imojie}
Keshav Kolluru, Samarth Aggarwal, Vipul Rathore, {Mausam}, and Soumen
  Chakrabarti. 2020{\natexlab{b}}.
\newblock \href {https://doi.org/10.18653/v1/2020.acl-main.521} {{IM}o{JIE}:
  Iterative memory-based joint open information extraction}.
\newblock In \emph{Proceedings of the 58th Annual Meeting of the Association
  for Computational Linguistics}, pages 5871--5886, Online. Association for
  Computational Linguistics.

\bibitem[{Li et~al.(2018{\natexlab{a}})Li, Yang, Song, and
  Hospedales}]{li2018learning}
Da~Li, Yongxin Yang, Yi{-}Zhe Song, and Timothy~M. Hospedales.
  2018{\natexlab{a}}.
\newblock \href
  {https://www.aaai.org/ocs/index.php/AAAI/AAAI18/paper/view/16067} {Learning
  to generalize: Meta-learning for domain generalization}.
\newblock In \emph{Proceedings of the Thirty-Second {AAAI} Conference on
  Artificial Intelligence, New Orleans, Louisiana, USA, February 2-7, 2018},
  pages 3490--3497. {AAAI} Press.

\bibitem[{Li et~al.(2018{\natexlab{b}})Li, Tian, Gong, Liu, Liu, Zhang, and
  Tao}]{li2018deep}
Ya~Li, Xinmei Tian, Mingming Gong, Yajing Liu, Tongliang Liu, Kun Zhang, and
  Dacheng Tao. 2018{\natexlab{b}}.
\newblock Deep domain generalization via conditional invariant adversarial
  networks.
\newblock In \emph{Proceedings of the European Conference on Computer Vision
  (ECCV)}, pages 624--639.

\bibitem[{Lu et~al.(2022)Lu, Liu, Dai, Xiao, Lin, Han, Sun, and
  Wu}]{lu2022unified}
Yaojie Lu, Qing Liu, Dai Dai, Xinyan Xiao, Hongyu Lin, Xianpei Han, Le~Sun, and
  Hua Wu. 2022.
\newblock Unified structure generation for universal information extraction.
\newblock In \emph{Proceedings of the 60th Annual Meeting of the Association
  for Computational Linguistics}, pages 5755--5772.

\bibitem[{Mi et~al.(2021)Mi, Ren, Dai, He, Sun, Li, Zheng, and
  Xu}]{mi2021towards}
Haitao Mi, Qiyu Ren, Yinpei Dai, Yifan He, Jian Sun, Yongbin Li, Jing Zheng,
  and Peng Xu. 2021.
\newblock Towards generalized models for beyond domain api task-oriented
  dialogue.
\newblock In \emph{AAAI-21 DSTC9 Workshop}.

\bibitem[{Muandet et~al.(2013)Muandet, Balduzzi, and
  Sch{\"{o}}lkopf}]{muandet2013domain}
Krikamol Muandet, David Balduzzi, and Bernhard Sch{\"{o}}lkopf. 2013.
\newblock \href {http://proceedings.mlr.press/v28/muandet13.html} {Domain
  generalization via invariant feature representation}.
\newblock In \emph{Proceedings of the 30th International Conference on Machine
  Learning, {ICML} 2013, Atlanta, GA, USA, 16-21 June 2013}, volume~28 of
  \emph{{JMLR} Workshop and Conference Proceedings}, pages 10--18. JMLR.org.

\bibitem[{Nan et~al.(2021)Nan, Zeng, Qiao, Guo, and Lu}]{nan2021uncovering}
Guoshun Nan, Jiaqi Zeng, Rui Qiao, Zhijiang Guo, and Wei Lu. 2021.
\newblock \href {https://doi.org/10.18653/v1/2021.emnlp-main.763} {Uncovering
  main causalities for long-tailed information extraction}.
\newblock In \emph{Proceedings of the 2021 Conference on Empirical Methods in
  Natural Language Processing}, pages 9683--9695, Online and Punta Cana,
  Dominican Republic. Association for Computational Linguistics.

\bibitem[{Niklaus et~al.(2018)Niklaus, Cetto, Freitas, and
  Handschuh}]{niklaus2018survey}
Christina Niklaus, Matthias Cetto, Andr{\'e} Freitas, and Siegfried Handschuh.
  2018.
\newblock \href {https://aclanthology.org/C18-1326} {A survey on open
  information extraction}.
\newblock In \emph{Proceedings of the 27th International Conference on
  Computational Linguistics}, pages 3866--3878, Santa Fe, New Mexico, USA.
  Association for Computational Linguistics.

\bibitem[{Pandey et~al.(2021)Pandey, Raman, Varambally, and
  Ap}]{pandey2021generalization}
Prashant Pandey, Mrigank Raman, Sumanth Varambally, and Prathosh Ap. 2021.
\newblock Generalization on unseen domains via inference-time label-preserving
  target projections.
\newblock In \emph{Proceedings of the IEEE/CVF Conference on Computer Vision
  and Pattern Recognition}, pages 12924--12933.

\bibitem[{Rasmussen and Ghahramani(2000)}]{rasmussen2000occam}
Carl~Edward Rasmussen and Zoubin Ghahramani. 2000.
\newblock \href
  {https://proceedings.neurips.cc/paper/2000/hash/0950ca92a4dcf426067cfd2246bb5ff3-Abstract.html}
  {Occam's razor}.
\newblock In \emph{Advances in Neural Information Processing Systems 13, Papers
  from Neural Information Processing Systems {(NIPS)} 2000, Denver, CO, {USA}},
  pages 294--300. {MIT} Press.

\bibitem[{Roy et~al.(2019)Roy, Park, Lee, and Pan}]{roy2019supervising}
Arpita Roy, Youngja Park, Taesung Lee, and Shimei Pan. 2019.
\newblock \href {https://doi.org/10.18653/v1/D19-1067} {Supervising
  unsupervised open information extraction models}.
\newblock In \emph{Proceedings of the 2019 Conference on Empirical Methods in
  Natural Language Processing and the 9th International Joint Conference on
  Natural Language Processing (EMNLP-IJCNLP)}, pages 728--737, Hong Kong,
  China. Association for Computational Linguistics.

\bibitem[{Solawetz and Larson(2021)}]{solawetz2021lsoie}
Jacob Solawetz and Stefan Larson. 2021.
\newblock \href {https://doi.org/10.18653/v1/2021.eacl-main.222} {{LSOIE}: A
  large-scale dataset for supervised open information extraction}.
\newblock In \emph{Proceedings of the 16th Conference of the European Chapter
  of the Association for Computational Linguistics: Main Volume}, pages
  2595--2600, Online. Association for Computational Linguistics.

\bibitem[{Stanovsky and Dagan(2016)}]{stanovsky2016creating}
Gabriel Stanovsky and Ido Dagan. 2016.
\newblock \href {https://doi.org/10.18653/v1/D16-1252} {Creating a large
  benchmark for open information extraction}.
\newblock In \emph{Proceedings of the 2016 Conference on Empirical Methods in
  Natural Language Processing}, pages 2300--2305, Austin, Texas. Association
  for Computational Linguistics.

\bibitem[{Stanovsky et~al.(2018)Stanovsky, Michael, Zettlemoyer, and
  Dagan}]{stanovsky2018supervised}
Gabriel Stanovsky, Julian Michael, Luke Zettlemoyer, and Ido Dagan. 2018.
\newblock \href {https://doi.org/10.18653/v1/N18-1081} {Supervised open
  information extraction}.
\newblock In \emph{Proceedings of the 2018 Conference of the North {A}merican
  Chapter of the Association for Computational Linguistics: Human Language
  Technologies,}, pages 885--895, New Orleans, Louisiana. Association for
  Computational Linguistics.

\bibitem[{Sun and Saenko(2016)}]{sun2016deep}
Baochen Sun and Kate Saenko. 2016.
\newblock Deep coral: Correlation alignment for deep domain adaptation.
\newblock In \emph{European conference on computer vision}, pages 443--450.
  Springer.

\bibitem[{Sun et~al.(2018)Sun, Li, Wang, Fan, Feng, and Li}]{sun2018logician}
Mingming Sun, Xu~Li, Xin Wang, Miao Fan, Yue Feng, and Ping Li. 2018.
\newblock \href {https://doi.org/10.1145/3159652.3159712} {Logician: {A}
  unified end-to-end neural approach for open-domain information extraction}.
\newblock In \emph{Proceedings of the Eleventh {ACM} International Conference
  on Web Search and Data Mining, {WSDM} 2018, Marina Del Rey, CA, USA, February
  5-9, 2018}, pages 556--564. {ACM}.

\bibitem[{Tzeng et~al.(2014)Tzeng, Hoffman, Zhang, Saenko, and
  Darrell}]{tzeng2014deep}
Eric Tzeng, Judy Hoffman, Ning Zhang, Kate Saenko, and Trevor Darrell. 2014.
\newblock Deep domain confusion: Maximizing for domain invariance.
\newblock \emph{arXiv preprint arXiv:1412.3474}.

\bibitem[{Wang et~al.(2020)Wang, Yu, Zhang, Liu, Zhu, and
  Sun}]{wang2020tplinker}
Yucheng Wang, Bowen Yu, Yueyang Zhang, Tingwen Liu, Hongsong Zhu, and Limin
  Sun. 2020.
\newblock Tplinker: Single-stage joint extraction of entities and relations
  through token pair linking.
\newblock In \emph{Proceedings of the 28th International Conference on
  Computational Linguistics}, pages 1572--1582.

\bibitem[{Wang et~al.(2021)Wang, Yu, Zhu, Liu, Yu, and
  Sun}]{wang2021discontinuous}
Yucheng Wang, Bowen Yu, Hongsong Zhu, Tingwen Liu, Nan Yu, and Limin Sun. 2021.
\newblock Discontinuous named entity recognition as maximal clique discovery.
\newblock In \emph{Proceedings of the 59th Annual Meeting of the Association
  for Computational Linguistics and the 11th International Joint Conference on
  Natural Language Processing (Volume 1: Long Papers)}, pages 764--774.

\bibitem[{Yu et~al.(2021)Yu, Wang, Liu, Zhu, Sun, and Wang}]{yu2021maximal}
Bowen Yu, Yucheng Wang, Tingwen Liu, Hongsong Zhu, Limin Sun, and Bin Wang.
  2021.
\newblock \href {https://doi.org/10.18653/v1/2021.emnlp-main.764} {Maximal
  clique based non-autoregressive open information extraction}.
\newblock In \emph{Proceedings of the 2021 Conference on Empirical Methods in
  Natural Language Processing}, pages 9696--9706, Online and Punta Cana,
  Dominican Republic. Association for Computational Linguistics.

\bibitem[{Yu et~al.(2020)Yu, Zhang, Shu, Liu, Wang, Wang, and Li}]{yu2020joint}
Bowen Yu, Zhenyu Zhang, Xiaobo Shu, Tingwen Liu, Yubin Wang, Bin Wang, and
  Sujian Li. 2020.
\newblock Joint extraction of entities and relations based on a novel
  decomposition strategy.
\newblock In \emph{Proceedings of ECAI}, pages 2282--2289. IOS Press.

\bibitem[{Zhang et~al.(2021)Zhang, Deng, Kawaguchi, Ghorbani, and
  Zou}]{zhang2020does}
Linjun Zhang, Zhun Deng, Kenji Kawaguchi, Amirata Ghorbani, and James Zou.
  2021.
\newblock \href {https://openreview.net/forum?id=8yKEo06dKNo} {How does mixup
  help with robustness and generalization?}
\newblock In \emph{9th International Conference on Learning Representations,
  {ICLR} 2021, Virtual Event, Austria, May 3-7, 2021}. OpenReview.net.

\bibitem[{Zhang et~al.(2020)Zhang, He, Liu, Lim, and
  Bing}]{zhang2020unsupervised}
Yan Zhang, Ruidan He, Zuozhu Liu, Kwan~Hui Lim, and Lidong Bing. 2020.
\newblock \href {https://doi.org/10.18653/v1/2020.emnlp-main.124} {An
  unsupervised sentence embedding method by mutual information maximization}.
\newblock In \emph{Proceedings of the 2020 Conference on Empirical Methods in
  Natural Language Processing}, pages 1601--1610, Online. Association for
  Computational Linguistics.

\bibitem[{Zhang et~al.(2022)Zhang, Yu, Yu, Liu, Fu, Li, Tang, Sun, and
  Li}]{zhang2022layout}
Zhenyu Zhang, Bowen Yu, Haiyang Yu, Tingwen Liu, Cheng Fu, Jingyang Li,
  Chengguang Tang, Jian Sun, and Yongbin Li. 2022.
\newblock Layout-aware information extraction for document-grounded dialogue:
  Dataset, method and demonstration.
\newblock In \emph{Proceedings of the 30th ACM International Conference on
  Multimedia}, pages 7252--7260.

\bibitem[{Zhao et~al.(2022)Zhao, Tian, Yao, Zheng, Lee, Song, Sun, and
  Zhang}]{zhao2022improving}
Yingxiu Zhao, Zhiliang Tian, Huaxiu Yao, Yinhe Zheng, Dongkyu Lee, Yiping Song,
  Jian Sun, and Nevin Zhang. 2022.
\newblock Improving meta-learning for low-resource text classification and
  generation via memory imitation.
\newblock In \emph{Proceedings of the 60th Annual Meeting of the Association
  for Computational Linguistics (Volume 1: Long Papers)}, pages 583--595.

\end{thebibliography}

\clearpage
\appendix

\section{GLOBE Dataset}
\label{sec:apd_dataset}

\subsection{Dataset Construction}
\label{sec:dataset}

To build GLOBE, we select six distinct data sources  for human annotation:
(1) \textbf{Insurance}, we use \begin{CJK}{UTF8}{gbsn}
保险条款
\end{CJK} (insurance policy) as the query, and retrieve relevant pdf documents in Baidu search engine\footnote{\url{https://www.baidu.com}} as the data source of the insurance domain;
(2) \textbf{Education}, we select the pages under the education topic in the Wikipedia classification index\footnote{\url{https://zh.wikipedia.org/wiki/}} as the data source of the education domain;
(3) \textbf{Finance}, we crawl public financial reports, including the stock market, business, investment, and other topics, as the data source of the finance domain;
(4) \textbf{Government}, we download official documents issued by government departments from the policy document library of the State Council\footnote{\url{http://www.gov.cn/zhengce/zhengcewenjianku/index.htm}} as the data source of the government domain;
(5) \textbf{Medicine}, we leverage the medical entity dictionary as a set of queries, and searched relevant texts in medical forums\footnote{\url{https://www.dxy.cn}} and online treatment manuals\footnote{\url{https://www.msdmanuals.cn/home}} as the data source of the medicine domain;
(6) \textbf{News}, we crawl news under the international news section of the Chinese News Service\footnote{\url{https://www.chinanews.com.cn/world/}} as the data source of the news domain.
We used PDFPlumber\footnote{\url{https://github.com/jsvine/pdfplumber}} to extract text from PDF documents, and used goose3\footnote{\url{https://github.com/goose3/goose3}} to extract the text of web pages.

We carefully select experienced annotators for dataset construction.
A principled training procedure is adopted to ensure the annotators are well trained, and the annotators are required to pass test tasks. 
All annotators are required to study the annotation guidelines of SAOKE carefully.
Before annotating GLOBE, the annotators need to have a test: labeling the sentences randomly selected in SAOKE and comparing them with the original annotations. 
Only those with a Gestalt F1 score greater than 0.95 are qualified for the final annotation.
Two annotators label each sentence, and if they have disagreements on one sentence, one or more annotators are asked to judge it.

\begin{table}[!h]
\normalsize
\centering
\resizebox{\linewidth}{!}{
\begin{tabular}{lcccccc}
\toprule
      & Ins  & Edu & Fin & Gov &  Med & News \\ \midrule
Number & 2,485  & 3,464  & 2,097  & 3,620 & 5,411 & 3822      \\
Percentage   & 11.9\%  & 16.6\%    & 10.0\%  & 17.3\% & 25.9\% & 18.3\%   \\  \bottomrule
\end{tabular}
}
\caption{The number and proportion of sentences belonging to different domains in GLOBE.}
\label{tab:domain_num}
\end{table}

\begin{table}[!h]
\normalsize
\centering
\resizebox{\linewidth}{!}{
\begin{tabular}{lcccc}
\toprule
      & Overlapping  & Discontinuous & Nested  & Complicated \\ \midrule
Number & 17,413  & 17,361  & 13,153  & 19,977       \\
Percentage   & 83.3\%  & 83.1\%    & 62.9\%  & 95.6\%   \\  \bottomrule
\end{tabular}
}
\caption{The number and proportion of sentences containing complicated facts in GLOBE.}
\label{tab:complicated_num}
\end{table}

\begin{table}[!h]
\normalsize
\centering
\resizebox{\linewidth}{!}{
\begin{tabular}{lccccc}
\toprule
    & [0,3]  & [4,6] & [7,9] & [10,12] & [13,$\infty$]  \\ \midrule
Number & 8,975  & 6,771  & 2,562  & 1,323 & 1,268       \\
Percentage   & 42.9\%  & 32.4\%    & 12.3\%  & 6.3\% & 6.1\%   \\  \bottomrule
\end{tabular}
}
\caption{The number and proportion of sentences containing different number of facts in GLOBE.}
\label{tab:fact_num_dis}
\end{table}

\subsection{Dataset Statistics}
The final GLOBE dataset consists 110,122 open facts annotated on 20,899 sentences spanning 6 distinct domains, making it the largest and most diverse human-annotated OpenIE test set.
This new dataset allows us to quantify the OpenIE performance in various downstream applications, and to better understand the limits of generalization exhibited by the most recent OpenIE methodology.
Table~\ref{tab:domain_num} shows the number and proportion of sentences belonging to different domains.
It can be found that there are at least 2k sentences in each domain, so the performance of OpenIE model can be fully measured.
We count the number of sentences in the data set that contains at least one complicated fact, as shown in Table~\ref{tab:complicated_num}. 
Here discontinuous means that at least one fact element in the sentence is not a continuous span, overlapping means that multiple facts in the sentence share at least one element, while nested means that different elements share some common spans.
It can be seen that identifying the discontinuous, overlapping, and nested facts is very important for OpenIE, because the sentences containing complicated facts account for 95.6\% in GLOBE.
We also report the fact number distribution in Table~\ref{tab:fact_num_dis}.
Most sentences contain more than 4 facts, and even 6.1\% sentences contain more than 12 facts, which increases the difficulty of extraction.
As presented in the detailed analysis part of the main experiment, our proposed DragonIE model attains consistent gains in complicated fact extraction and multiple fact extraction.


%

%

%


\begin{figure*}[!ht]
\centering
\subfigure[Complicated Extraction]{
\includegraphics[width=0.31\linewidth]{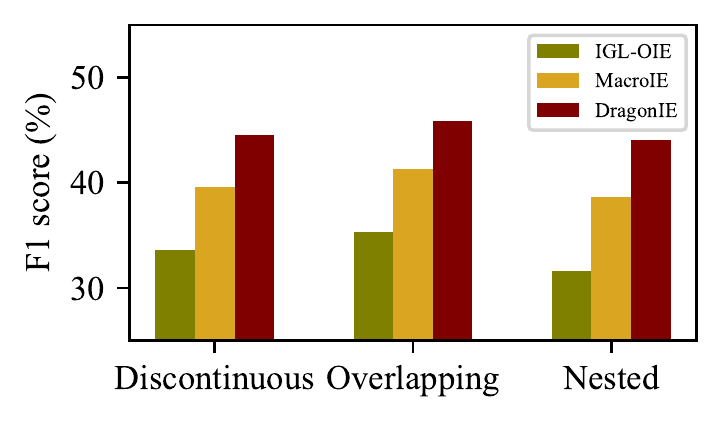}
}
\subfigure[Multiple Extraction]{
\includegraphics[width=0.31\linewidth]{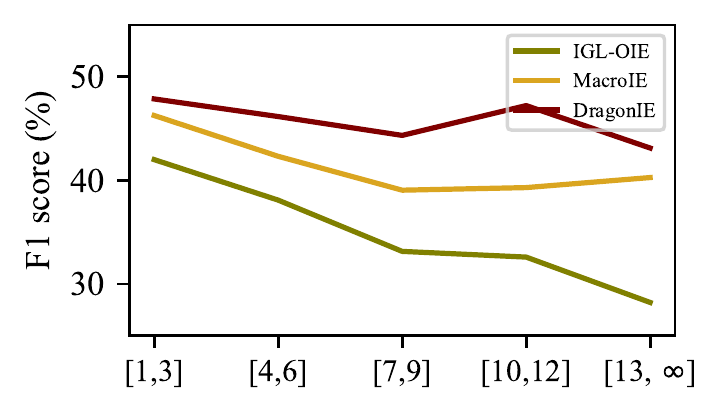}
}
\subfigure[Low-resource Extraction]{
\includegraphics[width=0.31\linewidth]{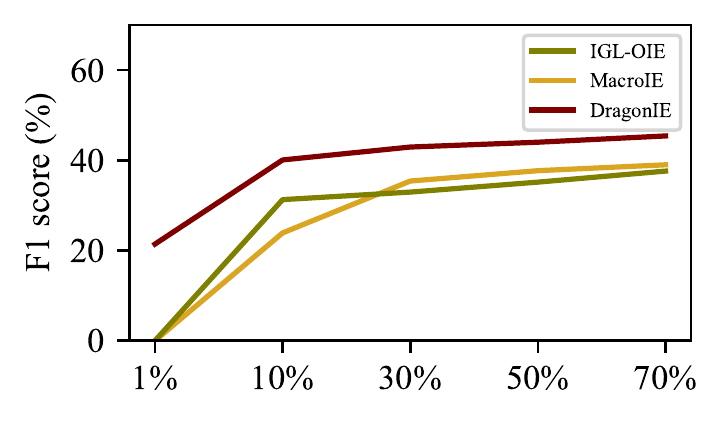}
}
\caption{Gestalt F1 scores on (a) complicated extraction, (b) multiple extraction, and (c) low-resource extraction. All the analyses are conducted on the SAOKE test set.}
\vspace{-1.0em}
\end{figure*}



\section{Detailed Experiments}

\subsection{Detailed Evaluation metrics}
We report performance values computed by the three most widely adopted metrics in the OpenIE literature.: (1) CaRB-single considers the number of common words in (gold, predicted) pair for each argument of the fact by greedily matching gold with one of the predicted facts; (2) CaRB-multi allows a gold fact to be matched to multiple predicted ones, thus more relaxed than CaRB-single; (3) Gestalt converts each fact into a string and uses the Gestalt function to measure the string similarity of (gold, predicted) pair. Therefore, it requires not only the coincidence of tokens, but also the consistency of token order, thus being the most stringent metric.

\subsection{Detailed Performance Comparison}
\label{sec:detail_comp}

Table~\ref{tab:globe-insurance}-\ref{tab:globe-news} summarize the detailed results in 6 domains of the GLOBE dataset. 
DragonIE has significantly exceeded the baseline model in 54 evaluation metrics of 6 domains, which once again proves the effectiveness of our method.
It is worth noting that there are great differences in the extraction performance in different domains, the highest F1 score of DragonIE is only 33.6\%, indicating that there is still much room for improvement toward practical out-of-domain applications.

\subsection{Detailed Analysis on SAOKE}
\label{sec:detail_saoke}

Similar with the detailed analysis conducted on GLOBE in the main experiment, we also perform a fine-grained evaluation on SAOKE.
(1) We select the sentences containing discontinuous or overlapping, or nested facts from SAOKE to form three complex test sets.
(2) We validate DragonIE's capability in extracting different numbers of open facts by splitting the sentences into five classes according to the fact count.
(3) We conduct low-resource experiments on five different partitions of the original SAOKE training sets (1/10/30/50/70\%).
As presented in Figure~\ref{fig:complicated}, DragonIE again attains gains in all classes across three settings,  consistent with the observation on GLOBE.

\subsection{Deatiled Analysis on Edge Type Space}
\label{sec:detail_edge}

In Table 2, we list the edge type sets of MacroIE and DragonIE on SAOKE (also GLOBE).
MacroIE needs 176 edge types, while DragonIE has only 21 edge types, reducing the edge types by 88\%.
Next, let's analyze the reasons carefully.
Theoretically, MacroIE needs $O(r^2) $ edge types, while DragonIE is $O(r) $, $r$ represents the number of possible role types in open facts.
There are 6 roles in SAOKE: \{\verb|subject|, \verb|predicate|, \verb|object|, \verb|time|, \verb|place|, \verb|qualifier|\}.

For MacroIE, different spans belonging to the same fact are connected to each other, by linking the beginning position and ending position of two spans, that is, there are 4 position types \{\verb|B2B|, \verb|B2E|, \verb|E2B|, \verb|E2E|\}.
There is also a \verb|NEXT| edge between adjacent spans belonging to the same kind of element to indicate the original order of spans.
Therefore, a total of $(6\times6+1)\times4=148$ edge types are required to represent the relations between 6 kinds of spans.
In addition, SAOKE also defines 7 virtual predicates\{ \verb|=|, \verb|BIRTH|, \verb|DEATH|, \verb|NOT|, \verb|DESC|, \verb|ISA|, \verb|IN|\}, which do not appear in the text. 
It is necessary to set virtual nodes for them and connect them to the boundary tokens of other elements in the fact. 
Therefore, $7\times4=28$ edge types are also required.
So MacroIE needs $148+28=176$ kinds of edges.

For DragonIE, it needs to set up a \verb|EB| type edge and a \verb|BE| type edge for each role, as well as a \verb|EE| edge and a \verb|I| edge.
To identify the virtual predicate, DragonIE connects the object to the virtual predicate node, so there are 7 additional edges.
So DragonIE needs $2\times6+2+7=21$ kinds of edges.

%

\begin{table*}[htb]
\centering {\footnotesize
\begin{tabular}{lcccccccrcc} 
\toprule       
Model $\downarrow$ - Metric $\rightarrow$ & \multicolumn{3}{c}{CaRB-single} & \multicolumn{3}{c}{CaRB-multi} & \multicolumn{3}{c}{Gestalt}
        \\ \cmidrule(r){2-4} \cmidrule(r){5-7} \cmidrule(r){8-10}
      & F1 & AUC & Opt.F1 & F1 & AUC & Opt.F1 & F1 & AUC & Opt.F1   \\
       
\midrule                         
IGL-OIE
& 18.1 & 5.6 & 18.7 & 21.3 & 7.7 & 22.2 & 15.0 & 4.0 & 14.0 \\
MacroIE
& 16.7 & 4.6 & 16.9 & 18.8 & 6.0 & 19.0 & 13.2 & 2.9 & 13.3 \\
\midrule 
DragonIE (ours)
& \textbf{24.7} & \textbf{9.5} & \textbf{25.3} & \textbf{28.1} & \textbf{12.5} & \textbf{29.0} & \textbf{20.8} & \textbf{7.4} & \textbf{21.4} \\
\bottomrule
\end{tabular}
\caption{\textbf{Out-of-domain Evaluation:} Main results on the \textbf{insurance} domain of GLOBE.
}
\label{tab:globe-insurance}
}
\end{table*}

\begin{table*}[htb]
\centering {\footnotesize
\begin{tabular}{lcccccccrcc} 
\toprule       
Model $\downarrow$ - Metric $\rightarrow$ & \multicolumn{3}{c}{CaRB-single} & \multicolumn{3}{c}{CaRB-multi} & \multicolumn{3}{c}{Gestalt}
        \\ \cmidrule(r){2-4} \cmidrule(r){5-7} \cmidrule(r){8-10}
      & F1 & AUC & Opt.F1 & F1 & AUC & Opt.F1 & F1 & AUC & Opt.F1   \\
       
\midrule                         
IGL-OIE
& 30.7 & 15.2 & 30.9 & 33.2 & 17.9 & 33.6 & 28.0 & 13.6 & 29.1 \\
MacroIE
& 31.0 & 13.3 & 31.0 & 32.7 & 15.0 & 32.7 & 28.4 & 10.9 & 28.4 \\
\midrule 
DragonIE (ours)
& \textbf{34.5} & \textbf{18.9} & \textbf{34.8} & \textbf{37.0} & \textbf{21.7} & \textbf{37.3} & \textbf{33.4} & \textbf{17.8} & \textbf{33.6} \\
\bottomrule
\end{tabular}
\caption{\textbf{Out-of-domain Evaluation:} Main results on the \textbf{education} domain of GLOBE.
}
\label{tab:globe-education}
}
\end{table*}

\begin{table*}[htb]
\centering {\footnotesize
\begin{tabular}{lcccccccrcc} 
\toprule       
Model $\downarrow$ - Metric $\rightarrow$ & \multicolumn{3}{c}{CaRB-single} & \multicolumn{3}{c}{CaRB-multi} & \multicolumn{3}{c}{Gestalt}
        \\ \cmidrule(r){2-4} \cmidrule(r){5-7} \cmidrule(r){8-10}
      & F1 & AUC & Opt.F1 & F1 & AUC & Opt.F1 & F1 & AUC & Opt.F1   \\
       
\midrule                         
IGL-OIE
& 22.2 & 8.6 & 22.6 & 24.5 & 10.4 & 25.0 & 19.0 & 6.2 & 19.5 \\
MacroIE
& 23.8 & 8.6 & 23.8 & 25.3 & 9.8 & 25.4 & 21.4 & 6.5 & 21.4 \\
\midrule 
DragonIE (ours)
& \textbf{30.1} & \textbf{13.5} & \textbf{30.1} & \textbf{32.6} & \textbf{15.7} & \textbf{32.7} & \textbf{26.9} & \textbf{11.0} & \textbf{27.2} \\
\bottomrule
\end{tabular}
\caption{\textbf{Out-of-domain Evaluation:} Main results on the \textbf{finance} domain of GLOBE.
}
\label{tab:globe-finance}
}
\end{table*}

\begin{table*}[htb]
\centering {\footnotesize
\begin{tabular}{lcccccccrcc} 
\toprule       
Model $\downarrow$ - Metric $\rightarrow$ & \multicolumn{3}{c}{CaRB-single} & \multicolumn{3}{c}{CaRB-multi} & \multicolumn{3}{c}{Gestalt}
        \\ \cmidrule(r){2-4} \cmidrule(r){5-7} \cmidrule(r){8-10}
      & F1 & AUC & Opt.F1 & F1 & AUC & Opt.F1 & F1 & AUC & Opt.F1   \\
       
\midrule                         
IGL-OIE
& 26.3 & 11.1 & 26.4 & 28.9 & 13.2 & 28.9 & 24.8 & 10.2 & 25.3 \\
MacroIE
& 28.3 & 12.5 & 28.3 & 30.1 & 14.3 & 30.3 & 26.1 & 10.2 & 26.2 \\
\midrule 
DragonIE (ours)
& \textbf{32.6} & \textbf{16.5} & \textbf{32.7} & \textbf{35.1} & \textbf{19.2} & \textbf{35.4} & \textbf{32.9} & \textbf{16.2} & \textbf{33.0} \\
\bottomrule
\end{tabular}
\caption{\textbf{Out-of-domain Evaluation:} Main results on the \textbf{government} domain of GLOBE.
}
\label{tab:globe-government}
}
\end{table*}

\begin{table*}[htb]
\centering {\footnotesize
\begin{tabular}{lcccccccrcc} 
\toprule       
Model $\downarrow$ - Metric $\rightarrow$ & \multicolumn{3}{c}{CaRB-single} & \multicolumn{3}{c}{CaRB-multi} & \multicolumn{3}{c}{Gestalt}
        \\ \cmidrule(r){2-4} \cmidrule(r){5-7} \cmidrule(r){8-10}
      & F1 & AUC & Opt.F1 & F1 & AUC & Opt.F1 & F1 & AUC & Opt.F1   \\
       
\midrule                         
IGL-OIE
& 26.6 & 12.0 & 26.7 & 29.0 & 13.9 & 29.1 & 21.0 & 8.5 & 21.3 \\
MacroIE
& 27.9 & 12.0 & 28.2 & 29.1 & 13.2 & 29.5 & 23.8 & 8.8 & 24.0 \\
\midrule 
DragonIE (ours)
& \textbf{34.5} & \textbf{18.5} & \textbf{34.6} & \textbf{36.4} & \textbf{20.6} & \textbf{36.6} & \textbf{31.0} & \textbf{15.0} & \textbf{31.1} \\
\bottomrule
\end{tabular}
\caption{\textbf{Out-of-domain Evaluation:} Main results on the \textbf{medicine} domain of GLOBE.
}
\label{tab:globe-medicine}
}
\end{table*}

\begin{table*}[htb]
\centering {\footnotesize
\begin{tabular}{lcccccccrcc} 
\toprule       
Model $\downarrow$ - Metric $\rightarrow$ & \multicolumn{3}{c}{CaRB-single} & \multicolumn{3}{c}{CaRB-multi} & \multicolumn{3}{c}{Gestalt}
        \\ \cmidrule(r){2-4} \cmidrule(r){5-7} \cmidrule(r){8-10}
      & F1 & AUC & Opt.F1 & F1 & AUC & Opt.F1 & F1 & AUC & Opt.F1   \\
       
\midrule                         
IGL-OIE
& 21.5 & 7.4 & 21.7 & 23.9 & 9.1 & 24.2 & 16.8 & 4.9 & 17.3 \\
MacroIE
& 20.1 & 5.9 & 20.1 & 21.6 & 6.9 & 21.7 & 17.2 & 4.1 & 17.2 \\
\midrule 
DragonIE (ours)
& \textbf{23.6} & \textbf{9.1} & \textbf{23.7} & \textbf{25.9} & \textbf{10.6} & \textbf{26.0} & \textbf{20.7} & \textbf{7.2} & \textbf{20.8} \\
\bottomrule
\end{tabular}
\caption{\textbf{Out-of-domain Evaluation:} Main results on the \textbf{news} domain of GLOBE.
}
\label{tab:globe-news}
}
\end{table*}

\begin{table*}[ht]
\centering
\small
\resizebox{0.9\textwidth}{!}{%
\begin{tabular}{p{0.1\textwidth}|p{0.9\textwidth}}
\toprule
Model & Edge Type Set \\ 
\midrule
MacroIE &
  ROLE-PAIR->predicate->qualifier-E2B, PREDEFINED-CLI->NOT-E2E, ROLE-PAIR->place->predicate-B2E, PREDEFINED-CLI->BIRTH-B2B, ROLE-PAIR->place->time-E2B, ROLE-PAIR->qualifier->time
  E2B, ROLE-PAIR->subject->qualifier-E2B, ROLE-PAIR->qualifier->predicate-E2E, ROLE-PAIR->subject->subject-E2E, ROLE-PAIR->subject->time-E2E, ROLE-PAIR->object->qualifier-E2B, ROLE-PAIR->time->object-E2B, PREDEFINED-CLI->DEATH-B2E, ROLE-PAIR->predicate->qualifier-B2B, ROLE-PAIR->object->qualifier-E2E, ROLE-PAIR->place->subject-E2E, ROLE-PAIR->time->place-B2B, ROLE-PAIR->subject->object-B2E, ROLE-PAIR->time->subject-E2E, PREDEFINED-CLI->NOT-B2E, ROLE-PAIR->place->subject-B2B, ROLE-PAIR->subject->qualifier-
E2E, ROLE-PAIR->object->subject-E2B, ROLE-PAIR->predicate->place-E2B, ROLE-PAIR->subject->time-B2E, ROLE-PAIR->subject->place-B2B, ROLE-PAIR->place->object-B2E, ROLE-PAIR->time->time-B2B, ROLE-PAIR->object->place-E2B, ROLE-PAIR->qualifier->predicate-B2B, PREDEFINED-CLI->ISA-B2E, ROLE-PAIR->time->predicate-E2E, ROLE-PAIR->time->time-E2B, ROLE-PAIR
  place->object-E2E, ROLE-PAIR->place->qualifier-B2B, PREDEFINED-CLI->=-B2E, ROLE-PAIR->qualifier->time-E2E, ROLE-PAIR->subject->place-B2E, ROLE-PAIR->subject->predicate-B2E, ROLE-PAIR->qualifier->object-E2E, ROLE-PAIR->predicate->place-B2B, PREDEFINED-CLI->DESC-B2B, ROLE-PAIR->place->predicate-E2E, ROLE-PAIR->subject->object-E2B, ROLE-PAIR->predicate
  qualifier-B2E, ROLE-PAIR->object->predicate-E2B, NEXT-E2E, ROLE-PAIR->time->qualifier-B2B, PREDEFINED-CLI->BIRTH-E2E, ROLE-PAIR->predicate->predicate-B2E, ROLE-PAIR->time->
predicate-E2B, ROLE-PAIR->object->place-B2B, ROLE-PAIR->qualifier->object-E2B, ROLE-PAIR->time->qualifier-E2E, ROLE-PAIR->subject->predicate-B2B, ROLE-PAIR->place->qualifier-E2B, ROLE-PAIR->place->place-B2E, ROLE-PAIR->time->predicate-B2E, ROLE-PAIR->subject->object-B2B, ROLE-PAIR->predicate->time-E2E, PREDEFINED-CLI->=-B2B, PREDEFINED-CLI->NOT-B2B, ROLE-PAIR->predicate->subject-E2E, ROLE-PAIR->qualifier->object-B2E, ROLE-PAIR->time->qualifier-E2B, ROLE-PAIR->time->time-E2E, ROLE-PAIR->place->predicate-B2B, ROLE-PAIR->object->subject-B2B, ROLE-PAIR->subject->qualifier-B2E, PREDEFINED-CLI->DEATH-B2B, ROLE-PAIR->object->place-B2E, ROLE-PAIR->object->predicate-B2B, PREDEFINED-CLI->IN
  B2E, ROLE-PAIR->object->subject-E2E, ROLE-PAIR->qualifier->subject-E2E, ROLE-PAIR->time->object-E2E, PREDEFINED-CLI->=-E2E, ROLE-PAIR->qualifier->predicate-E2B, ROLE-PAIR->subject->time-B2B, ROLE-PAIR->time->object-B2E, ROLE-PAIR->time->subject-B2E, ROLE-PAIR->subject->subject-B2B, ROLE-PAIR->object->qualifier-B2B, PREDEFINED-CLI->BIRTH-E2B, ROLE-PAIR->object->object-B2B, ROLE-PAIR->subject->predicate-E2B, ROLE-PAIR->qualifier->qualifier-E2B, ROLE-PAIR->subject->place-E2E, ROLE-PAIR->object->time-E2B, ROLE-PAIR->time->predicate-B2B, ROLE-PAIR->object->qualifier-B2E, ROLE-PAIR->predicate->time-E2B, ROLE-PAIR->time->qualifier-B2E, ROLE-PAIR->predicate->object-B2B, ROLE-PAIR->place->time-B2E, ROLE-PAIR->predicate->qualifier-E2E, PREDEFINED-CLI->DESC-E2E, ROLE-PAIR->predicate->predicate-E2B, ROLE-PAIR->qualifier->predicate-B2E, ROLE-PAIR->predicate->place-B2E, ROLE-PAIR->object->object-B2E, ROLE-PAIR->qualifier->place-B2E, PREDEFINED-CLI->DESC-B2E, ROLE-PAIR->time->place-B2E, ROLE-PAIR->subject->predicate-E2E, PREDEFINED-CLI
  ISA-E2E, ROLE-PAIR->object->predicate-E2E, ROLE-PAIR->predicate->object-E2B, PREDEFINED-CLI->DEATH-E2E, PREDEFINED-CLI->IN-E2E, ROLE-PAIR->qualifier->qualifier
  B2E, ROLE-PAIR->object->time-B2E, ROLE-PAIR->object->subject-B2E, ROLE-PAIR->place->subject-B2E, NEXT-B2E, PREDEFINED-CLI->NOT-E2B, NEXT-B2B, ROLE-PAIR->place->object-E2B, ROLE-PAIR->predicate->subject-E2B, PREDEFINED-CLI->DESC-E2B, PREDEFINED-CLI->IN-E2B, ROLE-PAIR->subject->subject-E2B, ROLE-PAIR->qualifier->subject-B2E, PREDEFINED-CLI->DEATH-E2B, ROLE-PAIR->object->time-E2E, PREDEFINED-CLI->ISA-B2B, ROLE-PAIR->place->subject-E2B, ROLE-PAIR->place->place-E2B, ROLE-PAIR->
subject->qualifier-B2B, ROLE-PAIR->qualifier->place-E2B, ROLE-PAIR->object->predicate-B2E, ROLE-PAIR->qualifier->place-B2B, ROLE-PAIR->qualifier->qualifier-B2B, ROLE-PAIR->predicate->subject-B2E, ROLE-PAIR->predicate->time-B2E, PREDEFINED-CLI->BIRTH-B2E, ROLE-PAIR->predicate->time-B2B, ROLE-PAIR->qualifier->place-E2E, ROLE-PAIR->qualifier->qualifier-E2E, ROLE-PAIR->time->subject-E2B, ROLE-PAIR->object->object-E2E, ROLE-PAIR->qualifier->subject-B2B, PREDEFINED-CLI->=-E2B, ROLE-PAIR->place->qualifier-B2E, ROLE-PAIR->predicate->predicate-B2B, ROLE-PAIR->subject->time-E2B, ROLE-PAIR->subject->object-E2E, ROLE-PAIR->place->qualifier-E2E, ROLE-PAIR->subject->place-E2B, ROLE-PAIR->predicate->
object-B2E, ROLE-PAIR->subject->subject-B2E, ROLE-PAIR->time->place-E2B, ROLE-PAIR->place->place-B2B, ROLE-PAIR->time->time-B2E, ROLE-PAIR->object->object-E2B, ROLE-PAIR->time->place-E2E, ROLE-PAIR->place->time-B2B, ROLE-PAIR->time->subject-B2B, PREDEFINED-CLI->IN-B2B, ROLE-PAIR->predicate->predicate-E2E, ROLE-PAIR->predicate->place-E2E, ROLE-PAIR->place->time-E2E, ROLE-PAIR->object->place-E2E, ROLE-PAIR->qualifier->time-B2B, ROLE-PAIR->object->time-B2B, NEXT-E2B, ROLE-PAIR->time->object-B2B, ROLE-PAIR->predicate->object-E2E, PREDEFINED-CLI->ISA-E2B, ROLE-PAIR->qualifier->object-B2B, ROLE-PAIR->qualifier->subject-E2B, ROLE-PAIR->predicate->subject-B2B, ROLE-PAIR->place->place
  E2E, ROLE-PAIR->qualifier->time-B2E, ROLE-PAIR->place->predicate-E2B, ROLE-PAIR->place->object-B2B
   \\ 
   \midrule  
DragonIE & 
  BE-object, BE-place, BE-predicate, BE-qualifier, BE-subject, BE-time, object->=, object->
BIRTH, object->
IN, object->NOT, object->DESC, object->ISA, object->DEATH, EE, I, EB-object, EB-place, EB-predicate, EB-qualifier, EB-subject, EB-time \\ \bottomrule
\end{tabular}%
}
\caption{The edge type set of MacroIE and DragonIE}
\label{tab:graph_space}
\end{table*}


\end{document}